\documentclass[letterpaper, 10 pt, conference]{ieeeconf}
\pdfoutput=1

\usepackage{times}
\usepackage{amssymb,amsfonts,amsmath,amscd}
\usepackage{enumerate}
\usepackage[usenames,dvipsnames]{xcolor}
\usepackage[
            hmargin={16.9mm,16.9mm},
            vmargin={20.1mm,15.2mm},
            footskip=0mm,
            includeheadfoot%
           ]{geometry}
\usepackage{url}
\usepackage{csquotes}
\usepackage{xspace}
\usepackage{MnSymbol}
\usepackage{array}
\usepackage{multirow}
\usepackage{listings} 
\usepackage{todonotes}
\usepackage{comment}
\usepackage{prettyref}

\makeatletter
\let\NAT@parse\undefined
\makeatother

\definecolor{Gray}{rgb}{0.3,0.3,0.3}
\usepackage[pdftex,
            breaklinks=true,
            colorlinks=true,
            citecolor=gray,
            linkcolor=Gray,
            urlcolor=blue
           ]{hyperref}
\usepackage{doi}

\newtoggle{compact}
\togglefalse{compact} 

\newtoggle{compactdiff}
\togglefalse{compactdiff} 

\newtoggle{todos}
\togglefalse{todos} 

\newcommand{\ignore}[1]{}

\iftoggle{compact}{%
    \newcommand{\compactversion}[1]{#1}
    \newcommand{\longversion}[1]{}
    \newcommand{\longversiontab}[1]{}
    \newcommand{\longref}[1]{}
    \newcommand{\compactversionStrict}[1]{#1}
    \newcommand{\longversionStrict}[1]{}
}{%
    \iftoggle{compactdiff}{
        \newcommand{\compactversion}[1]{\color{red}{}#1\color{black}{}}
        \newcommand{\longversion}[1]{{\color{OliveGreen}{}#1}}
        \newcommand{\longversiontab}[1]{\color{green}{}#1\color{black}{}}
        \newcommand{\longref}[1]{{\color{green}{}#1}}
    }{
        \newcommand{\compactversion}[1]{}
        \newcommand{\longversion}[1]{#1}
        \newcommand{\longversiontab}[1]{#1}
        \newcommand{\longref}[1]{#1}
    }
    \newcommand{\compactversionStrict}[1]{}
    \newcommand{\longversionStrict}[1]{#1}
}

\iftoggle{todos}{%
    \newcommand{\hannes}[1]{\todo[inline,color=green]{\small{TODO for Hannes: #1}}}
    \newcommand{\michael}[1]{\todo[inline,color=cyan]{\small{TODO for Michael: #1}}}
    \newcommand{\igor}[1]{\todo[inline,color=blue]{\small{TODO for Igor: #1}}}
    \newcommand{\discuss}[1]{\todo[inline,color=yellow!40]{\textbf{Discuss:} #1}}
}{%
    \newcommand{\hannes}[1]{}
    \newcommand{\michael}[1]{}
    \newcommand{\igor}[1]{}
    \newcommand{\discuss}[1]{}
}

\newcommand{\construction}[1]{\todo[inline,color=red!40]{\textbf{THIS IS A CONSTRUCTION SITE:} #1}\xspace}

\newrefformat{sec}{\ref{#1}}
\newrefformat{app}{App.~\ref{#1}}
\newrefformat{par}{Paragraph~\ref{#1}}
\newrefformat{hc}{HQMC\ref{#1}}
\newrefformat{reason}{\ref{#1})}
\newrefformat{fig}{Fig.~\ref{#1}}
\newrefformat{tab}{Table~\ref{#1}}
\newrefformat{row}{row~\ref{#1}}
\newrefformat{mex}{Example~\ref{#1}}

\def\quotebig#1{\textquote{\it #1}}
\def\quoteinline#1{\textquote{\it #1}}

\newcommand{\highlight}[1]{{\color{red}{\ensuremath{#1}}}}
\newcommand{\good}[1]{{\color{green}{\ensuremath{#1}}}}
\newcommand{\different}[1]{{\color{blue}{\ensuremath{#1}}}}

\title{Why and How to Avoid the Flipped Quaternion Multiplication}

\IEEEoverridecommandlockouts                              

\author{Hannes Sommer$^{\ast}$, Igor Gilitschenski$^{\sharp}$, Michael Bloesch$^{\dagger}$, Stephan Weiss$^{\ddagger}$, Roland Siegwart$^{\ast}$,\\and Juan Nieto$^{\ast}$
\thanks{%
The authors are affiliated with, the
$^{\ast}$Autonomous Systems Lab, ETH, Z\"{u}rich, CH,
$^{\sharp}$Computer Science and Artificial Intelligence Laboratory, MIT, Cambridge, MA, USA,
$^{\dagger}$Dyson Robotics Lab, Imperial College London, UK,
$^{\ddagger}$Institute of Smart System Technologies, Alpen-Adria-Universitat Klagenfurt, AT;
       {e-mail: \tt\small hannes.sommer@mavt.ethz.ch}%
}
}%

\newcommand{\bbm}{\begin{bmatrix}}
\newcommand{\ebm}{\end{bmatrix}}

\DeclareMathAlphabet{\mbf}{OT1}{ptm}{b}{n}
\newcommand{\mbs}[1]{{\boldsymbol{#1}}}

\DeclareMathAlphabet{\mathbfit}{OML}{cmm}{b}{it}

\def\N{\mathbb N}
\def\R{\mathbb R}
\def\H{\mathbb H}

\newenvironment{notation}{\vspace{3mm}\begin{tabular}[t]{r @{\;\;\;:\;\;\;} p{.80\textwidth}}}{\end{tabular}\vspace{3mm}}

\DeclareMathOperator{\sign}{sign}

\begin{document}
\setcounter{tocdepth}{2}
\maketitle

\begin{abstract}
Over the last decades quaternions have become a crucial and very successful tool for attitude representation in robotics and aerospace.
However, there is a major problem that is continuously causing trouble in practice when it comes to exchanging formulas or implementations: there are two quaternion multiplications in common use, Hamilton's multiplication and its flipped version, which is often associated with NASA's Jet Propulsion Laboratory.
We believe that this particular issue is completely avoidable and only exists today due to a lack of understanding.
This paper explains the underlying problem for the popular passive world to body usage of rotation quaternions, 
and derives an alternative solution compatible with Hamilton's multiplication. 
Furthermore, it argues for entirely discontinuing the flipped multiplication.
Additionally, it provides recipes for efficiently detecting relevant conventions and migrating formulas or algorithms between them.
\end{abstract}

\longversion{\tableofcontents}

\newcommand{\p}{\mbf p}
\newcommand{\q}{\mbf q}
\newcommand{\x}{\mbf x}
\newcommand{\C}{\mbf C}
\newcommand{\CS}{\C_S}
\newcommand{\CH}{\C_H}
\newcommand{\avel}{\mbs\omega}

\section*{Nomenclature}
\newcommand\imquatvec[1]{\overrightarrow{\mathbf{#1}}}%

\newcommand{\crossMat}[1]{{{#1}^\times}}%
\newcommand{\qid}{\mbf 1}
\newcommand{\imc}[1]{{\mathcal{I}(#1)}}%
\newcommand{\xv}{\imquatvec{\x}}%
\newcommand{\qv}{\imquatvec{\q}}%
\newcommand{\pv}{\imquatvec{\p}}%
\newcommand{\imcinv}[1]{{\mathcal{I}^*(#1)}}
\newcommand{\SOm}{{\operatorname{SO}(3)}}
\newcommand{\SOGm}{{\operatorname{SO}}}

\def\rotV{{\mbs \phi}}%
\def\U{\mathcal U}%
\renewcommand{\notation}[2]{{#1}&{#2}\\}%
\begin{flushleft}
\setlength\tabcolsep{1pt}

\begin{tabular}{@{}p{0.145\columnwidth} p{0.84\columnwidth}@{}}
\longversiontab{
\notation{$\x, \mbf M$}{\emph{vectors} and \emph{matrices} in upper case bold letters.
    {\bf If not specified otherwise matrices are always assumed to be in $\R^{3\times 3}$.}
}}
{$\R^{k\times 1}$}&{$\simeq\R^k$; \textbf{implicitly} identified $\forall k\in\N$}\\
\notation{$\cdot \times \cdot$}{\emph{cross product}}
\notation{$\crossMat{\cdot}$}{$:\R^3 \to \R^{3\times 3}, \forall \x, \mbf y\in \R^3$ : $(\crossMat{\x})\mbf y = {\x} \times \mbf y$}
\notation{$\H$}{
	$:=(\R^4, +, \odot)$, \emph{quaternions}, $(\qid, \mbf i, \mbf j, \mbf k):=(\mbf e_i)_{i=1}^4$
}
\notation{$\U$}{
    $:=\{\q\in \H \, | \, \|\q\| = 1\}$, \emph{unit length} quaternions
}
\notation{\compactversion{$\odot\footnotemark{}$}\longversiontab{$\p\q, \cdot, \odot\footnotemark{}$}}{
Hamilton's multiplication, i.e. $\longversiontab{\mbf i\mbf j=\mbf i\cdot \mbf j=}\mbf i\odot \mbf j = \mbf k$
}
\notation{$\otimes$}{
flipped mult. \cite{shuster93}: $\forall \p,\q \in \H : \q\otimes\p = \p\odot\q$
}
\notation{$\imquatvec{\q}\!,\imc{\q}$}{$:\H \to \R^3, \q=(q_1, q_2, q_3, q_4) \mapsto (q_2, q_3, q_4)$}
\notation{$\imcinv{\x}$}{$:\R^3 \to \H, \x \mapsto x_1 \mbf i + x_2\mbf j + x_3 \mbf k \Rightarrow \mathcal I \circ \mathcal{I}^{*} = \text{id}_{\R^3}$}
\longversiontab{
\notation{$\q\odot\x, \newline \q\otimes\x$}{both quaternion mult. are assumed overloaded for the pairs in $\R^3 \times \R^4$ or $\R^4 \times \R^3$, such that the $\R^3$ element is treated as purely imaginary quaternion:
e.g.  $\forall \q\in \R^4, \x \in \R^3: \x \otimes \q := \q\odot\x := \q \odot \imcinv{\x}$
}%
}%
\notation{$\bar\q$}{quaternion \emph{conjugation}, $\bar \q := (q_1, -q_2, -q_3, -q_4)$}
\notation{$\SOm$}{$:= (\{ \mbf M \in \R^{3\times 3} \, | \, \mbf M^{-1} = \mbf M^T \wedge \det(\mbf M) = 1 \}, \cdot)$}
\end{tabular}
\end{flushleft}
\footnotetext{$\odot$ is only used here for better readability (to contrast $\otimes$). For documents using Hamilton's multiplication only we recommend to just use the normal $\cdot$ or the default operator (omitted symbol), because it is the default multiplication symbol for quaternions in mathematics, which should not be needlessly questioned as the standardizing authority.}

\section{Introduction}

\quotebig{The quaternion \cite{shuster93} is one of the most important representations of the attitude in spacecraft attitude estimation and control.}
\longversion{
}%
With these words Malcolm D. Shuster opened his introduction of \cite{shuster2008}, ``The nature of the quaternion'' (in 2008). 
It details on a conventional shift from Hamilton's original quaternion multiplication, $\odot$, to its flipped version, $\otimes$, he had successfully advocated for in the 1990s - in particular with the very influential \cite{shuster93}.
There, the flipped multiplication was called \emph{natural}%
\footnote{In robotics this multiplication is often and inaccurately referred to as \emph{JPL-convention}.
While this convention uses Shuster's multiplication, it also includes additional conventional decisions.}
and the original was called \emph{traditional}.
Because we consider ``natural'' too tendentious and because both labels are generally rather unknown we will refer to the former with Shuster's multiplication and to the latter with Hamilton's multiplication.
\longversion{
}%
Back then\longref{\footnote{See \prettyref{app:history} for more historical details.}}, the goal of the introduction was to prevent confusion and mistakes when dealing with chains of rotations (\emph{compositions}) motivated by the application of spacecraft attitude estimation and control \cite{shuster2008}.
However, the introduced conventional split with other fields (e.g. mathematics, physics, or computer graphics / visualization) caused a lot of confusion and was a source for potential mistakes in practice at least in other fields, such as robotics.
Therefore, in this paper, we advocate to undo this split by only using Hamilton's original multiplication.
At the same time we show how an alternative solution can be used to solve the original problem addressed in \cite{shuster93}.

\subsection{Original problem and Shuster's solution}
The \emph{historical problem} with Hamilton's multiplication appeared only together with an arguably arbitrary, specific convention on how to assign direction cosine matrices (DCM)\footnote{Equivalent in value to a corresponding coordinate transformation matrix (between oriented orthonormal bases).}, $\CS : \U\to \SOm$ to unit length quaternions. 
When using this assignment, as in \cite{shuster93}, the product of two quaternions, $\p \odot \q$, would correspond to the product of the two corresponding DCMs, but with reversed order:
\begin{equation}
\label{eq:problem} 
\forall \p,\q\in\U : \CS(\p \odot \q) = \CS(\q)\cdot \CS(\p).
\end{equation}
I.e. the mapping, $\CS$, was not a \emph{homomorphism} $(\U, \odot)\rightarrow(\SOm, \cdot)$ but an \emph{antihomomorphism}, which is generally more surprising and therefore more error-prone\longversion{ and less convenient}.
Introducing the flipped $\otimes$ is \emph{Shuster's solution} to this problem \cite{shuster93}:
\begin{equation}
\p \otimes \q:=\q \odot \p \label{eq:shuster-product}
\;\; \Rightarrow \;\; 
\CS(\p \otimes \q) = \CS(\p)\cdot \CS(\q).
\end{equation}

\subsection{The problem today with Shuster's solution}
\label{sec:problem}
Since its introduction, a great fraction of spacecraft literature switched to $\otimes$ \cite{shuster2008}.
On the other hand virtually the entire rest of the scientific community dealing with quaternions is apparently not following, and partially not even aware of this transition in the spacecraft\longversion{ estimation and control} community as we will show in \prettyref{sec:lit-review}.
Having two different quaternion multiplications in use comes at a significant and continuous cost: formulas and implementations\longversion{ from the two different conventions} need translation or adaptation let alone one first must identify which multiplication is used in a given formula or algorithm.
The latter can be particularly tedious if the authors are not aware of the two possibilities and hence do not mention which one they use%
\footnote{As a small self experiment, the reader in doubt is invited to try to figure out which quaternion multiplication the popular c++ template library for linear algebra, Eigen, is using.\longref{ Our solution is in \prettyref{app:libEigen}}}%
.
If a reader is unaware of the split, the discovery that two different quaternion multiplications are in use, and that in fact ``the other'' was employed, might be made only after several failures\longversion{, during which the confusion may even have spread to third parties}.
\longversion{This was already well observed in \cite{sola2012quaternion}.}%

The troubles may be little per subject but they affected and will affect a significant number of scientist and engineers causing delays in the best case.
\longversion{
}%
This accumulated cost is particularly large for interdisciplinary domains such as robotics that lean on publications from both conventional worlds, for instance aerospace and computer vision.
In fact, in robotics there are several recent publications for each of the two multiplications (see \prettyref{sec:lit-review}).
It is also a pressing matter because the effort to correct becomes larger with every new implementation or publication using a convention different from what the community agrees on eventually.

\subsection{Contribution}
Given this current situation, it is an interesting question whether the proposed switch was necessary\longversion{ or at least the best of all available options}.
Or phrased differently, whether some benefit is actually worth this conventional split or whether there is an economical alternative\longversion{ even when considering the effort already invested in the partial transition}. 
Addressing these questions this work provides the following contributions%
\compactversion{\footnote{An extended version of this paper with in-depth arguments, derivations, proofs, and more additional material can be found here \cite{sommer2017arXve}.}}%
:%
{%
\begin{itemize}
  \item identify the problem of two quaternion multiplications\longversion{ to improve awareness}
  \item investigate the necessity of the flip \eqref{eq:shuster-product} to solve \eqref{eq:problem}
  \item advertise and explain one alternative solution
  \item demonstrate that this alternative yields more formal similarity to corresponding formulas using matrices
  \item advocate to discontinue Shuster's multiplication\longversion{ in favour of Hamilton's original definition}
  \item provide recipes to detect and migrate between quaternion multiplication conventions
\end{itemize}
}%

\noindent To the best of our knowledge, this is the first work explicitly addressing the ambiguity of quaternion multiplication as a problem and proposing a potential solution.
\longversion{Our work allows for reducing the overall cost associated with having too many competing conventions about the same thing.}

\subsection{Outline}
\longversion{\hannes{FINAL:Adapt to final structure}}
We start with important background information in \prettyref{sec:usage}.
In \prettyref{sec:lit-review}, we summarize relevant contributions on quaternion multiplication related conventions, 
and tabulate their popularity in recent or influential\longversion{\footnote{regarding rotation quaternion convention}} publications.
Next, we propose an alternative solution in \prettyref{sec:alternative-solution}.
Then we argue for discontinuing Shuster's multiplication in \prettyref{sec:rationale}.
Lastly, we provide recipes for detecting and migrating rotation quaternion related conventions in \prettyref{sec:recipes}.
\longversion{Further supplemental information is provided in the appendix.}

\section{Background: Duality of rotation representations}\label{sec:usage}
Physical rotations are conceptually rather unambiguous\longversion{\footnote{For example as rigid body motions keeping at least one point fixed.}}.
However, when it comes to describing them, there are two popular ways to do so.
The \emph{active} way is to describe what happens to the rotated body's points from a fixed \emph{world frame}'s perspective.
The \emph{passive} way is to describe how coordinates of fixed world points change from the perspective of the rotated \emph{body frame}\footnote{The two ways are also distinguished as \emph{alibi} (active) and \emph{alias}.}.
It comes in two flavours, \emph{world-to-body} (PWTB) and \emph{body-to-world} (PBTW).
The first describes the change from before rotation to after and the second from after to before%
\footnote{The names stem from assuming that body and world frame are initially aligned. Alternative: local to global vs. global to local \cite{sola2012quaternion}.}%
.
We call these options the \emph{usage}\footnote{\longversion{There don't seem to be established names. }\cite{sola2012quaternion} uses \quoteinline{function} for active vs. passive without flavours.} of the mathematical model to describe the physical rotation.
Mathematically, the described actions on the observed points have identical properties in all three cases matching precisely the actions of the special orthogonal group ($\SOGm$\footnote{This paper only deals with $\SOm$.})\longversion{ on the underlying Euclidean space}.
But because given the \emph{same} physical rotation, PWTB describes the inverse compared to PBTW, they yield mutually inverse group elements.
Despite of being conceptually very different, the active usage yields always the same group element as PBTW.
Consequentially we will often not distinguish these two.\longversion{ See \prettyref{app:rotation_matrices} for an in-depth discussion of the various ways to use matrices to represent physical rotations.

}%
The crucial consequence is that while every \emph{parametrization} of the $\SOGm$ can be employed for all usages, switching usage from or to PWTB corresponds to an inversion.
\longversion{We believe that a lack of awareness for this fact ultimately led to the \quoteinline{problem} \eqref{eq:problem} (see \prettyref{sec:big-picture-comparison}).
}%
The main target audience of this paper are those who favour the PWTB usage because the problem \eqref{eq:problem} is exclusive for the PWTB usage.

\section{Literature review}
\label{sec:lit-review}
\longversion{
Not many publications discuss the multitude of possible conventions or argue for a specific choice, instead the common way seems to either just assume a convention or define one more or less completely before using it.
}
\subsection{Literature on or introducing relevant conventions}
\longversion{\subsubsection{Aerospace}}
According to \cite{crassidis2006sigma} Shuster's multiplication, $\otimes$, was introduced in \cite{lefferts1982kalman}. 
It was later proposed as a standard in \cite{breckenridge1999} according to \cite{Trawny2005}.
Unfortunately, the exact details are impossible for us to reconstruct as NASA is not publishing \cite{breckenridge1999}.
In \cite{junkins86} the authors derive a homomorphic composition rule for \emph{Euler (symmetric) parameter} (ESP)\footnote{Usually treated as equal to the quaternion components} (effectively $\otimes$) and infer a required conjugation from ESP to Hamilton's quaternions.
In \prettyref{sec:alternative-solution} we are proposing the same quaternion values\longversion{ as alternative solution} but with different derivation.
\longversion{
According to \cite[p.473]{shuster93} (7 years later), the authors of \cite{junkins86} in fact preferred Shuster's multiplication over Hamilton's.
This indicates that they considered Euler parameter together with Shuster's multiplication a separate representation distinguished from Hamilton's quaternions by a conjugation.
}%

In \cite{shuster2008} the author describes the transition from Hamilton's $\odot$ to Shuster's $\otimes$ for \quoteinline{spacecraft attitude estimation and control} literature
\compactversion{
 citing \cite{shuster93} as \quoteinline{probably, more than any other work,} responsible for the conventional shift.
}\longversion{%
:
\quotebig{%
Hamilton's approach to quaternions \ldots, seemingly in universal use until the publication of reference \cite{Markley1978} and still in almost universal
use until the publication of reference \cite{shuster93}, which, probably, more than any other work, has been responsible for the change to the natural order of quaternion multiplication in spacecraft attitude estimation and control.
This was, in fact, an avowed purpose of the author of reference \cite{shuster93}.
But although nearly every writer on spacecraft attitude is aware now of reference \cite{shuster93}, which is cited frequently, he or she may not be aware of the inconsistency of reference \cite{shuster93} with many other works on quaternions.}%
}%
In \cite{shuster93} $\otimes$ is indeed derived as necessity given some requirements, of which one was the correspondence $\CS$.
\compactversion{%
At the same time Shuster speculated already why mathematicians have no reason to reform the quaternions:
}%
\longversion{%
In order to demonstrate a clear trend towards generally adopting the new multiplication order \cite{shuster2008} cites recent publications dealing with rotation quaternions: four using Shuster's multiplication and one using Hamilton's multiplication.
The fact that the first were all aerospace paper and the last a book on applied mathematics might be related to Shuster's observation in \cite{shuster93}, when explaining why mathematicians might not have changed the quaternion multiplication order:
}
\quoteinline{The concern of pure mathematics is not in representing physical reality efficiently but in exploring mathematical structures \ldots{} As engineers, our interest is in ``im-pure'' mathematics, contaminated by the needs of practical application.}
\longversion{
}%
A major struggle with the quaternion conventions within NASA is reported in \cite{yazell2009origins}, including the discussion of a major conventional switch from the official Space Shuttle program to the (American) International Space Station software standard.
The author compares these two conventions with a third convention that he calls \emph{Robotics}.
Neither \cite{shuster93} nor \cite{shuster2008} are mentioned in \cite{yazell2009origins} despite their apparent relevance. 
The only easily accessible reference in \cite{yazell2009origins} for quaternions, \cite{stevens2003aircraft}, uses Hamilton's multiplication and also does not mention Shuster's work.

The \emph{Navigation and Ancillary Information Facility} (NAIF), a sub organization of NASA's Jet Propulsion Laboratory (JPL), uses Hamilton's multiplication everywhere in their primary software suite called \emph{SPICE} \cite{naif2003}.
\longversion{%
In that document, the authors introduce two conventions: \emph{SPICE Quaternions}, using Hamilton's multiplication, and \emph{Alternate Style Quaternions}, using Shuster's multiplication.
They present important formulas in both \textquote{styles} and stress intensively how important it is to pay attention to the small difference in the details.
}%
The authors use the short terms \emph{SPICE quaternions} and \emph{Engineering quaternions} and describe the first with \quoteinline{Invented by Sir William Rowan Hamilton; Frequently used in mathematics and physics textbooks} and the second with \quoteinline{Widely used in aerospace engineering applications} (e.g.  \href{https://naif.jpl.nasa.gov/pub/naif/toolkit_docs/C/cspice/qxq_c.html}{cspice/qxq\_c}).

\subsubsection{Robotics}
In \cite{sola2012quaternion} the author proposes a classification of rotation quaternion related conventions based on four binary choices leading to  \quoteinline{12 different combinations}\footnote{12 (instead of 16) because one binary choice, body to world vs world to body only makes sense in the passive case.}.
The choice between homomorphic and anti-homomorphic matrix to quaternion conversion (see \prettyref{sec:theory}) is not among these choices. 
Instead homomorphy is implicitly assumed as a natural requirement.
The author also points out two most commonly used conventions, one using Shuster's and one using Hamilton's multiplication.
The author chooses to use Hamilton's multiplication because it 
\quotebig{%
... coincides with many software libraries of widespread use in robotics, such as Eigen, ROS, Google Ceres, and with a vast amount of literature on Kalman filtering for attitude estimation using IMUs (\cite{Chou92, kuipers1999quaternions, pinies07inertial, roussillon2011rt, martinelli2012vision}, and many others).%
}
At the same time PBTW usage is chosen, circumventing the problem \eqref{eq:problem}.

\subsection{Popularity in influential and recent literature}\label{sec:pupularity}
In \prettyref{tab:current-state} we group publications, which are recently dealing with rotation quaternions, by which quaternion multiplication is used, by the type of publication, and the scientific community for books and papers.
Of course, this is merely a small selection of corresponding publications in the respective fields.
\begin{table}[t]\footnotesize
\setlength\tabcolsep{2pt}
    \caption{Popularity of the two multiplications\protect\footnotemark{}}
    \centering
    \begin{tabular}{p{0.26\columnwidth}|p{0.45\columnwidth}|p{0.22\columnwidth}}
        Type/Community & Hamilton\protect\footnotemark{} & Shuster \\ \hline
        Online encyclopedia & 
        \href{http://www.encyclopediaofmath.org/index.php?title=Quaternion}{Encyclopedia of Math.}, 
        \href{http://mathworld.wolfram.com/Quaternion.html}{Wolfram Mathworld}, 
        \href{http://planetmath.org/hamiltonianquaternions}{Planetmath}, 
        \href{http://www.britannica.com/topic/quaternion}{Britannica}, 
        \href{http://en.wikipedia.org/wiki/Quaternion}{Wikipedia}
        & \\
        \hline
        Mathematics 
        & 
        \cite{kuipers1999quaternions}(B!),
        \cite{hazewinkel2013encyclopaedia}(Ba),
        \cite{gowers2010princeton}(Ba),
        \cite{lengyel2012mathematics}(Baw),
        \cite{rodman2014topics}(B-)
        & 
        \\
        \hline
        Aerospace 
        & 
        \cite{junkins86}(B),
        \cite{battin1987}(Bw),
        \cite{kuipers1999quaternions}(B!),
        \cite{stevens2003aircraft}(B!), 
        \cite{shivarama2004hamilton}(w),
        \cite{bacon2012quaternion}(!)
        & 
        \cite{lefferts1982kalman, shuster93, crassidis2006sigma, crassidis2007survey, Barfoot1100, ainscough2014q, queen2015kalman, galante2016fast, mccabe2017comparison} \\
        \hline
        Robotics 
        & 
        \cite{Pervin82}(a),
        \cite{Chou92}(aw),
        \cite{Murray9400}(Bw), 
        \cite{roussillon2011rt}(w), 
        \cite{martinelli2012vision}(w),\!
        \cite{sola2012quaternion}(w),
        \cite{fresk2013full}(w)
        & 
        \cite{Trawny2005},
        \cite{diebel2006representing},
        \cite{Barfoot1100},
        \cite{li2012improving},
        \cite{li2014high}
        \\
        \hline
        Mechanics 
        & 
        \cite{schwab2002quaternions}(w), 
        \cite{schwab2006draw}(w),
        \cite{holm2008geometric}(Baw),
        \cite{heard2008rigid}(Ba),
        \cite{flügge2013principles}(Ba) 
        \cite{2016mechanics}(Ba) 
        \\
        \hline
        Control 
        & 
        \cite{shivarama2004hamilton}(w), 
        \cite{bonnable2009invariant}(!),
        \cite{fresk2013full}(w),
        \cite{Haichao2016}(!)
        &
        \\
        \hline
        Computer vision
        & 
        \cite{Pervin82}(a),
        \cite{mukundan2002quaternions}(aw),
        \cite{roussillon2011rt}(w)
        &
        \\
        \hline
        Computer graphics / visualization &
        \cite{shoemake1987quaternion}(!),
        \cite{mukundan2002quaternions}(aw),
        \cite{hanson2005visualizing}(aw), 
        \cite{vince2011quaternions}(B!), 
        \cite{lengyel2012mathematics}(Baw),
        \cite{Kolaman2012}(-)
        & 
        \\
        \hline
        Applications and software libraries & 
        \href{http://reference.wolfram.com/language/Quaternions/tutorial/Quaternions.html}{Wolfram Mathematica},
        Matlab's \href{http://ch.mathworks.com/help/aerotbx/ug/quatmultiply.html}{aerospace}(!) and  
        \href{http://ch.mathworks.com/help/robotics/}{robotics toolbox}, 
        C++ library Eigen\longref{ (see \prettyref{app:libEigen})}, 
        \href{https://github.com/ceres-solver/ceres-solver/}{Google Ceres}, 
        \href{http://www.boost.org/doc/libs/1_64_0/libs/math/test/quaternion_test.cpp}{Boost}, 
        \href{https://octave.sourceforge.io/quaternion/overview.html}{GNU Octave}, 
        \href{http://docs.ros.org/jade/api/tf2/html/Quaternion_8h_source.html}{ROS}, 
        NASA's \href{https://naif.jpl.nasa.gov/pub/naif/toolkit_docs/C/cspice}{SPICE} 
         (\href{https://naif.jpl.nasa.gov/pub/naif/toolkit_docs/C/cspice/qxq_c.html}{qxq\_c},
         \href{https://naif.jpl.nasa.gov/pub/naif/toolkit_docs/C/cspice/m2q_c.html}{m2q\_c}) 
        & \href{https://msdn.microsoft.com/en-us/library/microsoft.directx_sdk.quaternion.xmquaternionmultiply(v=vs.85).aspx}{Microsoft's DirectXMath Library}\footnotemark
        \\
        \hline
    \end{tabular}
    \label{tab:current-state}
\end{table}
\addtocounter{footnote}{-2}%
\footnotetext{The references are roughly in chronological order.}%
\addtocounter{footnote}{1}%
\footnotetext{Parentheses decorate differences from a default (article, passive world-to-body usage):  (B) book, (-) no rotations used, (a) active usage, (w) passive body-to-world usage, (aw) $\simeq$ (a) or (w), (!) antihomomorphic QM-convention (i.e. using $\CS$ in the Hamilton column).}%
\addtocounter{footnote}{1}%
\footnotetext{This is an unclear case since the documentation of the multiplication function states: \textquote{Returns the product of two quaternions as Q2*Q1}, where the multiplication function arguments are in the order Q1, Q2. I.e. Shuster's multiplication is implemented while the documentation uses Hamilton's multiplication (implicitly).}%
Within these constraints the table shows:
All relevant online general purpose or mathematical encyclopedias, which we could find, use Hamilton's multiplication (without mentioning any alternatives).
And so do all major software packages.
Most of them also use precisely the quaternion to matrix mapping we suggest as part of an alternative solution to \eqref{eq:problem} in \prettyref{sec:alternative-solution}.
The only scientific communities we could find using Shuster's multiplication are aerospace and robotics. 
Both seem to be still divided.
Even NASA itself and it's substructure JPL seem to be divided to this day.

\section{Proposed alternative solution}
\label{sec:alternative-solution}
The alternative solution we suggest for the original problem \eqref{eq:problem} is to use a different correspondence rule, $\CH: \U\ni \q \mapsto \CS(\bar\q)$.
Assuming we keep fixed how rotation matrices represent rotations, this new correspondence inverts the corresponding quaternion.
It also yields the desired homomorphy from quaternions to rotation matrices together with $\odot$ and the usual matrix multiplication:
\begin{equation}
\label{eq:our-solution} 
\forall \p,\q\in\U : \CH(\p\odot \q) = \CH(\p)\cdot \CH(\q). 
\end{equation}
This suggestion might give the wrong negative impression of an ``additional'' inversion. 
But it could equally well be that for $\CS$ the ``wrong'' (inverted) quaternions were chosen.
In fact our observations in \prettyref{sec:big-picture-comparison} support this interpretation\longversion{ and allow it to be explained as follows: the mistake to use active quaternions in a passive usage scenario led to $\CS$, which in turn led to flipping the multiplication, while choosing appropriate passive quaternions would have lead to $\CH$ and no need to flip the multiplication}.
\longversion{

}%
Please note that both solutions yield the usual way to apply the rotation represented by a quaternion\footnote{This fact is actually equivalent to both being homomorphic.} (see \ref{mex:quatrot}):
\begin{equation}\label{eq:defQuatMatMaps}
\forall {\x \in\R^3, \q \in \U} : \begin{matrix}\CH(\q) \x = \imc{\q\odot \imcinv{\x}\odot \q^{-1}}\\ \CS(\q) \x = \imc{\q\otimes \imcinv{\x }\otimes \q^{-1}}\end{matrix}
\end{equation}

\longversion{
We are going to refer to the $\CH$ rotation quaternions with \emph{Hamilton's rotation quaternions} because they yield homomorphy together with Hamilton's quaternion multiplication (see \prettyref{sec:theory}).}

\begin{figure}
  \centering
    \includegraphics[]{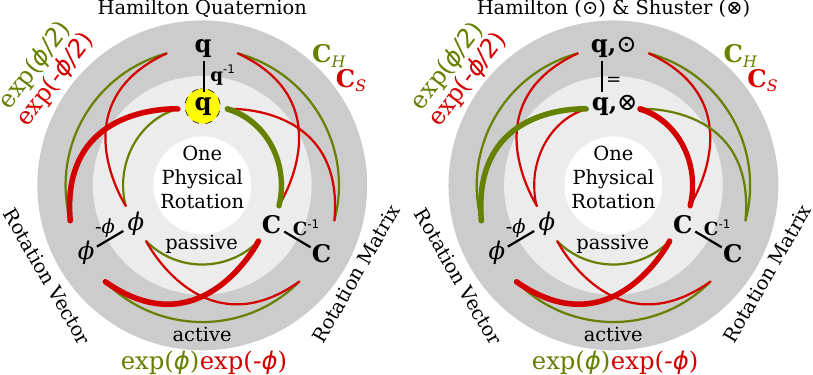}%
  \caption{%
  Active (or passive body-to-world; outer ring) and passive (world-to-body; inner ring) representatives of a single physical rotation and their mutual conversions: black lines convert only between active and passive, green convert between different parametrizations, and red additionally invert.
  The bold lines on the left (Hamilton multiplication only) constitute the homomorphic convention suggested in this paper.
  It is suitable for everybody preferring passive world-to-body for $\q$ and $\C$.
  \longversion{The rare users of passive rotation vectors would entirely remain in the inner circle.
  }%
  Clinging to the green conversion from the common active $\rotV$ to $\q$ even if the quaternion is intended to be employed passively, as done in $\cite{shuster93}$ leads to the situation on the right (using Shuster's multiplication when employing quaternions passively).
  This neglects the option of the highlighted passive $\q$ (on the left) and thus requires the flipped multiplication, $\otimes$, and inevitably breaks the symmetry between $\C$ and $\q$ on the right compared to the left.
  }
  \label{fig:rotation_conversion}
\vspace{-1em}
\end{figure}

\section{Rationale against Shuster's multiplication}
\label{sec:rationale}
\longversion{\subsection{Overview}}
First of all, it is important to emphasize that the question which quaternion multiplication to use commonly, should be considered as a pragmatic question.
Conceptually there is no problem with any of the two multiplications or with having two multiplications commonly used.
Having said that, our main argument is as follows:

\subsubsection{Both multiplications are equally capable and interchangeable for every application}\label{reason:both_are_equally_good}
As we explain in \prettyref{sec:equally-capable} one multiplication is as good as the other in terms of applicability. 
For example to represent rotations.
\longversion{There is a difference for those rare people who use the components of a rotation quaternion intuitively. However, for these we suggest a better alternative in \prettyref{sec:perspective_intuitive}.
Having both \quoteinline{available} at the same time does not add extra expressiveness because one can always be substituted by the other with flipped arguments.}

\subsubsection{One of the two multiplications should be discontinued}\label{reason:one_must_go}
Having two quaternion multiplications in formulas and software comes at a significant and ongoing cost (see \prettyref{sec:problem}, \prettyref{sec:lit-review}), but does not provide benefits (see \ref{reason:both_are_equally_good}).

\subsubsection{Hamilton's multiplication is preferable}\label{reason:hamilton_is_better}
Hamilton's multiplication is predominant over all existing publications.
This is because i) it is about 150 years older, and ii) Shuster's multiplication did not spread to mathematics or theoretical physics and probably will never do.
The latter arises from the fact that both disciplines have very different needs that typically do not even lead to the original problem \eqref{eq:problem} as already observed in \cite{shuster93}.
To the best of our knowledge it only spread from aerospace to parts of robotics, which has similar needs and is strongly influenced\longversion{ (among others)} by aerospace literature.
Also, Hamilton's multiplication is used by the vast majority of current software\longversion{ (libraries)} (see \prettyref{tab:current-state}).
\longversion{Interestingly this includes the SPICE package developed at NAIF for and as part of the very same JPL at which Shuster's multiplication was proposed as a standard \cite{breckenridge1999}. }%
These observations make it extremely difficult to discontinue Hamilton's multiplication instead.
Even more so, pure mathematics should be almost impossible to influence in that respect because it is against the very nature of mathematics to change such a definition without an inner-mathematical reason%
.
\longversion{

}%
Additionally, as we show in \prettyref{sec:comparison}, Hamilton's multiplication has its benefits over Shuster's from a formal point of view.
These benefits make it less error prone if combined with\longversion{ the alternative quaternion to rotation matrix mapping,} $\CH$, which is at least as old and well-known\footnote{$\CH$ is\longversion{ identical to the one given by the well-known formula, \eqref{eq:eulerRodrigues}, } often referred to with \emph{Euler-Rodrigues formula}\longversion{, as demonstrated in \prettyref{app:eulerRodrigues}}.} as $\CS$.
It is typically used whenever rotations are represented actively or PBTW (see \prettyref{sec:usage}).
Because of that, $\CH$ is also already predominant among existing software implementations (see \prettyref{tab:current-state}).

Employing $\CH$ also for the very popular PWTB usage, as we suggest, has a consequence that one could consider its biggest drawback:
It introduces the comparatively rarely seen \emph{passive quaternion} (see \prettyref{tab:comparison}, \prettyref{fig:rotation_conversion}), and thus induces an unusual conversion from an active rotation vector, $\rotV$: $\q(\rotV) = \exp(- 0.5 \rotV)$.
Eventually, the original problem $\eqref{eq:problem}$ leaves two options: i) introducing an unusual quaternion multiplication (Shuster's solution) or ii) introducing an unusual conversion from the active rotation vector to the (passive) rotation quaternion (used e.g. in \cite{junkins86}).
Having only these two options is probably the very difficulty behind the overall struggle and what prevented a standardization so far (see \prettyref{sec:lit-review}). 
\longversion{Also, general purpose text books do not help resolving this issue as they typically do consider neither i) nor ii) by either accepting antihomomorphy or only consider active / PBTW usage (see \prettyref{sec:pupularity}).

}%
Our position, in favour of ii), emerges from rephrasing the question equivalently: What should be specific for the passive usage of quaternions when compared to the active usage, their multiplication or the conversion to the active\footnote{There is good reason to call the usual rotation vector active. But it is not required here. Our argument works as well for neutral rotation vector.} rotation vector?
It reveals an immediate argument for ii): the conversion from active rotation vector to passive quaternion should even be expected to be different from its conversion to an active quaternion (compare also \prettyref{sec:usage}).
An argument against i) is that the multiplication is used much more often, because conversions are merely interface operations and usually do not appear where the heavy lifting is done.
Also, if we have to chose the multiplication depending on the usage of quaternions to represent rotations, which multiplication is right for other applications?
Furthermore, it is very unusual and therefore unexpected that a fundamental algebraic operation such as a multiplication is a matter of dispute and convention, while it is quite common for representations.
Finally, the very same question for rotation matrices was answered historically: there is no flipped matrix multiplication in regular use. 
Instead it is fully accepted that there is an active and a passive rotation matrix\longversion{ mutually inverse and} related differently to an active rotation vector.
Why not accept the exact same for rotation quaternions while employing them for exactly the same purpose?

\subsection{Equal capability argument}
\label{sec:equally-capable}
In this section we explain why both multiplications are equally capable to support \prettyref{reason:both_are_equally_good}.
The migration recipes in \prettyref{sec:migration} show that any algorithm or formula can be migrated between the two multiplications with very simple steps: just introduce quaternion conjugation at some places and exchange all multiplications.
This already shows that both multiplications can provide the same service to the user (writer, reader, programmer), 
as the migration concepts work in both directions.
\longversion{

}%
The question remains whether one implementation could be computationally more efficient than the other.
However, one should easily be convinced that the extra cost for some additional conjugations is negligible for most use cases since it only involves scalar negations.
Furthermore, for very performance critical parts of an implementation, the negations can almost always be statically merged into other operations or simplified away mathematically\longversion{ (compare also \prettyref{sec:perspective_algorithmic})}.

\subsection{Formal differences}
\label{sec:theory}
\newcommand{\SOa}{\mathcal{SO}(3)}

\newcommand{\M}{\C}
\newcommand{\MH}{\CH}
\newcommand{\MS}{\CS}

\def\hqmcH{$(\MH, \odot)$\xspace}
\def\hqmcS{$(\MS, \otimes)$\xspace}
\def\comp{\mathcal{C}}

\compactversion{
In this section we are going to investigate the formal differences of the two alternative solutions to the original problem \eqref{eq:problem} to support parts of \prettyref{reason:hamilton_is_better}.
Please note that this does not contradict our observation of equal capability in \prettyref{sec:equally-capable}, because formal similarity or symmetry do not affect the practical capability of a mathematical tool. 
Instead it affects mostly the probability of human error when dealing with the tool.
And this can be a strong argument, too. 
In fact it is the very type of argument with which \cite{shuster93} justifies the introduction of Shuster's multiplication.

We start without any assumptions about how unit quaternions and proper orthogonal matrices are conceptually used for modeling rotations\footnote{For instance active/passive/DCM matrices.\longref{ See also \prettyref{app:matrix-representation})}} and only focus on how quaternions are converted to matrices.
This conversion must be at least implicitly determined by all consistent rotation conventions involving rotation matrices and rotation quaternions.
The benefit is that our abstract observations apply automatically to all such conventions.
In particular the $\C$ in $\CS$ and $\CH$ does no longer stand for direction cosine matrix, but for any rotation matrix concept.

\subsubsection{QM-convention}
We call a convention that determines a specific conversion, $\C: \U \to \SOm$, and a particular quaternion multiplication, $\star$, a \emph{QM-convention} and represent it with ($\C, \star$).
In this way many common conventions collapse to a single representative. 
However, that is intended as their differences do not matter for the following analysis, i.e. it holds in any case.
We call a QM-convention \emph{homomorphic} iff the multiplication order is preserved when switching between quaternions and matrices, as in \eqref{eq:our-solution}, \emph{antihomomorphic} iff the order flips, as in \eqref{eq:problem}, and inconsistent otherwise.
There are only four options for consistent QM-conventions, namely all pairs in $\{\MH, \MS\} \times \{ \odot, \otimes\}$\longref{ (see \prettyref{sec:formal}), of which two are homomorphic and two are antihomomorphic}.
It seems commonly accepted that a homomorphy is preferable and less error-prone reducing the good candidates to the two \emph{homomorphic QM-convention}s\longversion{ (HQMC)}, \hqmcH and \hqmcS, with their homomorphy given in \eqref{eq:our-solution} and \eqref{eq:shuster-product} respectively.
}

\longversion{
One major problem when writing about representations for rotations is that there are many different representations and for most of them there are multiple conventions concerning their details and mutual relation (see e.g. \cite{shuster93}, \cite{sola2012quaternion}\longref{ or \prettyref{app:rotation_matrices}}).
In fact there is not even a reliable agreement on what precisely proper rotations are (e.g. active vs. passive rotations) or --- more generally --- how to distinguish the different things to represent.

For this section we choose a different approach: we restrict ourselves to the relation between unit quaternions and special orthogonal matrices while leaving open how the latter are used to represent physical rotations or similar.
This way the analysis stays general and less complex because most of the usual complexity comes from the how and what to represent.
This strategy is possible because there is a common aspect considering all usual ways to model proper physical rotations with matrices:
They formulate a one to one relationship with the special orthogonal $\R^{3 \times 3}$ matrices, $\SOm:=\{\M\in \R^{3\times3} | \M^T = \M, \det(\M) = 1\}$ with their group operation (matrix multiplication) corresponding to some composition of rotations --- just in different ways.
The direction cosine matrix used in the introduction as an example is only one possible interpretation of $\SOm$.
See \prettyref{app:matrix-representation} for a discussion of all the common interpretations.

This little common ground is already enough to formulate the original problem when restricting it to the relation between unit quaternions, $\U:=\{\q\in \H = (\R^4, +, \odot), \|\q\| = 1\}$, and $\SOm$ by entirely forgetting about their potential usage.
This relation has different meaning depending on which meaning is assigned to $\U$ and $\SOm$ but all common formal possibilities how to map $\U$ to $\SOm$ can be analyzed entirely without referring to this meaning.
This way a thorough formal comparison becomes more tractable.

Because of this possibility we will, after introducing some essential notions, adopt a purely formal perspective to compare the common different ways for how to map $\U$ to $\SOm$ and its relation to the quaternion multiplication and rephrase the original problem in that context.
This will leave us with two optimal solutions: the one that was suggested in \cite{shuster93} and the one we are proposing to adopt instead.
To compare them further we will investigate the relation to the cross product, the rotation vector and the angular velocity in \prettyref{sec:comparison} and additionally to the active usage in \prettyref{sec:big-picture-comparison}.
These comparison will show significant differences in favour of the solution including Hamilton's multiplication.
After that we are going to verify that non of the other perspectives, numeric, algorithmic and intuition give strong reasons for one of the two option over the other.
In particular, we will explain why non of them performs better from any of these other perspectives.
This way, overall, our suggestion will remains as preferable.

\subsection{Essential notions}
To support a concise communication in this \prettyref{sec:theory} it is important to clarify some essential notions, namely \emph{rotation} and \emph{convention} of rotation representation.

\subsubsection{Rotation}
For the sake of simplicity we refer to 3-dimensional proper rotations or attitude / orientation or coordinate transformations between orthonormal coordinate systems with just \emph{rotation}s. 
Despite the fact that these are quite different concepts this is possible because for this paper's cause and arguments their distinction is irrelevant.

\subsubsection{Partial rotation convention}
In order to allow reliable arguments for a claim like both multiplications are equally capable we need some rigorous notion to specify what precisely we compare.
As tools for rotation representation the quaternion multiplications do not stand alone.
Instead they need to be combined with rules such as how to use them to rotate a vector, which typically involve choosing one of multiple available options (see for example the binary choices of \cite{sola2012quaternion}).
A set of such rules / determinations is often referred to as \emph{convention}.
What we actually want to compare are classes of such conventions.
For instance the class of conventions using Hamilton's multiplication with the class using Shuster's multiplication.

One way to support this would be to define the convention notion in a rigorous way. 
But this is difficult mostly because it is not clear what specifications are necessary for a convention to be complete.
In practice it is very different how much authors actually specify when they declare the convention they use in a text.
Typically not more than needed for that text, which is highly dependent on the text.
This problem we avoid by introducing the new concept of a \emph{partial rotation convention} or shorter \emph{partial convention}.
They can play the role of convention classes without relying on a rigorous notion of a complete / full convention.

We refer to any set of arbitrary determinations concerning rotation representation, i.e. not necessary in that particular way, with \emph{partial rotation convention}.
These determinations can be explicit or implicit, i.e. logical / mathematical consequences from the explicit determinations.
This is quite analogous to the notion of an axiom system for a given domain.
Therefore we call the explicit determinations \emph{axioms}.
Examples of such determinations are conversion rules between rotation parametrizations, how a orthogonal matrix is interpreted as rotation, or how to multiply two rotation quaternions.
Fortunately it is not necessary for our cause to specify all kind of determinations that can be made by a partial convention.

\paragraph{Consistency}
A partial convention not contain contradicting axioms we call a \emph{consistent} partial convention.
One hypothetical non-consistent example could be when a convention includes all conversion rules in the triangle of unit quaternions, orthogonal matrices, and angle axis parameters but converting a quaternion first into angle axis and then into a matrix can yield a different result from when the same quaternion is directly converted to a matrix.
We only care for consistent conventions.

\paragraph{Natural partial order}
\label{sec:natural-convention-order}
These partial conventions naturally posses a partial order, just like sets induced by their inclusion.
We say a partial convention, $C$, is \emph{included} in another convention, $D$, ($C \subseteq D$), iff the union of their determinations would yield a consistent partial convention and this union would not determine anything more than $D$.
This fits well the intuitive expectation that when adding an axiom to a convention it becomes a larger convention.
In practice it is sufficient and necessary for $C\subseteq D$ that all axioms of $C$ are also determined by $D$.
This makes it particularly easy for a $C$ with only a few axioms.

\paragraph{QM-convention}
We call a partial convention that determines a map from unit quaternions to orthogonal matrices and a particular quaternion multiplication a \emph{QM-convention}.
We call a partial convention, $C$ a \emph{minimal QM-convention} iff it is a QM-convention and minimal with respect to their partial order, i.e. there is no partial convention $\hat C \subseteq C$ such that it isn't also $C \subseteq \hat C$.
Such a partial convention does not determine anything more than absolutely necessary for a QM-convention.
One important fact about partial conventions is that everything that can be inferred from the axioms of a given convention applies to all larger partial conventions.
This ultimately makes our investigation of minimal QM-conventions in \prettyref{sec:formal} relevant for the partial conventions used in practice, which are typically (much) larger.

\subsection{Mathematical / formal perspective}
\label{sec:formal}

Given a way to represent rotations with matrices there are different approaches to employ unit quaternions, $\U$, as representation for physical rotations.
One approach, the \emph{matrix-induced} method, is to parametrize these matrices with unit quaternions, by specifying a surjective mapping $\U \to \SOm$, and retrieve the induced rotation representation by the unit quaternions.
Others approaches are to model rotations directly with unit quaternions or specify their relation to other available rotations representations.
Once a unit quaternion representation is given the conversion to any other representation including any matrix representation can be formally retrieved.
Any such conversion from unit quaternion representation to a matrix representation would yield a surjective mapping $\U \to \SOm$.
It must be surjective because the matrices are one to one representations in all cases. 
I.e. leaving one out would mean missing a physical rotation.
Therefore one can find the the same representation applying the matrix-induced approach.

For this section we are adopting the matrix-induced approach because it involves a minimal number of choices while being at least as powerful.
We assume the choice for a specific matrix representation to be fixed but unknown to make sure the analysis does not depend on that choice.
We are going to compare a set of two mappings only, $\MH, \MS$, from $\U$ to $\SOm$, which can be uniquely defined by \eqref{eq:defQuatMatMaps}.
To the best of our knowledge, these two mappings constitute effectively all mappings that were found and successfully used in the history of representing physical rotations.

\subsubsection{Problem formalization}
By the means of both mappings the unit quaternions become singularity free parametrizations of $\SOm$, the \emph{Euler-Rodrigues parameters} in case of $M_1$.
The problem, \eqref{eq:problem}, comes with the comparison of matrix and quaternion multiplication, each inducing a Lie-group structure on the respective sets, through $M_1$ and $M_2$.
It turns out that the mappings $\MH,\MS$ are - from this algebraic perspective - smooth and surjective group \emph{homomorphisms} ($\MH$) and \emph{antihomomorphisms} ($\MS$) respectively from $\U$ to $\SOm$:
\begin{equation}\label{eq:m1m2}\begin{aligned}
\forall {\p,\q\in \U } : \MH(\p\q ) 
    &=\MH(\p) \MH(\q) \wedge \MS(\p\q ) \\
    &=\MS(\q) \MS(\p)
\end{aligned}\end{equation}
This property follows immediately from the defining property \eqref{eq:defQuatMatMaps}, because both multiplications (quaternion and matrix) are associative.

When switching from $(\U,\odot)$ to its \emph{opposite} group $\U^\text{op} = (\U, \otimes)$ with the flipped multiplication, as suggested by \cite{shuster93}, a very similar statement becomes true,
with the only difference that --- as in general when switching one group to its opposite group -- homomorphisms and antihomomorphisms switch places when kept fixed as mappings between the two sets, because the identity mapping between the two is an antiisomorphism.
\begin{equation}\label{eq:defQuatMatMapsOp}\begin{aligned}
\forall {\p,\q\in \U } : \MS(\p\otimes\q ) 
    &=\MS(\p) \MS(\q) \wedge \MH(\p\otimes\q ) \\
    &=\MH(\q) \MH(\p)
\end{aligned}\end{equation}
Because $\U$ and $\U^\text{op}$ are isomorphic, by means of 
\begin{equation}
\psi:\q\mapsto\q^{-1}
\end{equation}
and because $\MS = \MH\circ \Psi \wedge \MH = \MS\circ \Psi$ they also switch places with respect to property \eqref{eq:defQuatMatMaps}:
\begin{equation}\label{eq:m1m2op}
\forall {\x \in\R^3} : \MS(\q) \x = \q\otimes \x \otimes\q^{-1}\wedge \MH(\q) \x =\q^{-1}\otimes \x\otimes \q
\end{equation}

The historic problem \eqref{eq:problem} and action consequently taken can now be phrased as follows:
Experience in spacecraft attitude estimation and control showed that dealing with rotation quaternions that behaved antihomomorphic compared to the orthogonal matrices used for the \emph{same} purpose was error prone and unaesthetic.
It was antihomomorphic, because $\MS$ was the mapping used together with Hamilton's quaternion multiplication, as in the second part of \eqref{eq:m1m2}.
For some reason it was out of question to change the mapping from quaternion to matrix from $\MS$ to $\MH$.
And therefore, it was suggested to flip the quaternion multiplication order - the \emph{only} alternative left besides flipping the matrix multiplication order - to fix the antihomomorphy of \eqref{eq:problem}.

In this paper we are claiming that it would have been better to switch from $\MS$ to $\MH$ instead, see \prettyref{reason:hamilton_is_better}.

The four possible outcomes of the two binary choices, $(\M, \star) \in \{\MH, \MS\} \times \{ \odot, \otimes\}$, yield all the minimal \emph{QM-convention}s included in any of the usual convention used in practice.
Hence all considerations concerning those four applies to all convention practically used, only depending on which of these four are included\footnote{Only one of them can be included in a consistent partial convention.}.

From the purely formal perspective it is well understandable that the two \emph{homomorphic QM-convention}s (HQMC), \hqmcH and \hqmcS, are favourable over the antihomomorphic QM-conventions, $(\MS, \odot)$ and $(\MH, \otimes)$.
In \cite{sola2012quaternion} partial conventions including the antihomomorphic pairs are not even considered possible conventions, because there the choice of the multiplication determines the mapping from quaternions to matrices by assuming the homomorphy through \eqref{eq:quatToMatMap}, as beyond dispute --- and we can only agree with that.

It is also apparent that there is nothing to be gained by choosing one option over the other among the two homomorphic QM-conventions --- from this formal perspective.
Both fulfill the same and only relevant and desired properties for a homomorphic QM-convention, $(\M, \star) \in \{(\MH, \odot), (\MS, \otimes)\}$
\begin{align}
\M(\q) \x &= \imc {\q \star \x \star \q^{-1}} \label{eq:quatToMatMap}\\
\Rightarrow \M(\p\star \q) &= \M(\p)\M(\q) \label{eq:pairHom}
\end{align}
as observed in \eqref{eq:defQuatMatMaps}, \eqref{eq:defQuatMatMapsOp} and \eqref{eq:m1m2}, \eqref{eq:m1m2op}.

\paragraph{Extension to the entire quaternions algebra} Sometimes derivations or algorithms exploit the fact that the unit quaternions are embedded in the algebra and skew-field of quaternions, $\H$.
But even together with this embedding the two options are isomorphic and therefore equally capable.
This can be seen as follows. The mapping
\begin{equation}\label{eq:antiisomorphism}
\Psi: (\H, \otimes) \to (\H, \odot), \q\mapsto\bar\q,
\end{equation}
where $\bar\q$ denotes the quaternion conjugate of $\q$, is a skew-field isomorphism $(\H, \otimes) \to (\H, \odot)$.
This can be easily verified using the well-known antihomomorphy properties of quaternion conjugation.
Furthermore, $\Psi$ is an extension of $\psi$ onto $\H$ in the sense that restriction of $\Psi$ onto $\U$ is $\psi$.
Additionally the quaternion norm $\|\cdot\|$ defining $\U$ commutes with $\Psi$.
This implies that the embedding relation between $\U$ and $\H$ fully compatible with $\Psi$.
Please note that $\Psi$ does not keep the imaginary units $\mbf i, \mbf j, \mbf k$ invariant. 
Instead they get mapped to their negative.

\subsubsection{Seeking formal differences that allow for a preference}
So far the formal comparison could not yield any preference for one of the two HQMCs.
Therefore in this section we are going to discuss the effect of this choice on the relation to additional rotational concepts.
Namely the rotation vector and the angular velocity.
Including these concepts in the comparison (\prettyref{tab:comparison}) will finally reveal an important difference.
Why these two extra concepts? 
First they suffice to formulate our argument and second the reasoning in \cite{shuster93} for flipping the quaternion multiplication was also only based on no further concepts.

First we briefly introduce the two additional concepts as far as needed for the comparison.
While introducing them we will implicitly adopt conventions that not everybody would agree with as is true for all other possible decisions.
This keeps the formulas more simple. 
And to compensate for the lack of conventional neutrality we demonstrate in \prettyref{app:sim_proof} that the conclusions we draw from the comparison are in fact not depending on the these particular conventional choices.

\paragraph{Rotation vector parametrization}
Considering a \emph{rotation vector}, $\rotV \in \R^3$ (magnitude corresponds to magnitude of rotation (rad) and axis to the axis of rotation) gives rise to a conversion formulas from rotation vector to quaternion or matrix.
Again we fix the formal functional relation instead of referring to any interpretation: 
\begin{equation}
\label{eq:rotVec}
\C(\rotV) =: \exp(-\crossMat{\rotV}).
\end{equation}
In literature the name rotation vector isn't consistently used. 
Other names are Euler vector or angle-axis vector.
\def\u{\mbf{u}}

One compatible and very common interpretation is for example the direction cosine matrix (DCM), $\C(\rotV)$ from a coordinate frame $A$ to a frame $B$ and a rotation vector $\rotV$, expressed in $A$ and describing the rotation necessary to rotate the basis of $A$ to overlap with the basis of $B$ assuming a right hand rule interpretation of the axis direction.
The rotation vector parametrization is in closely related to the angle-axis parametrization, $(\alpha,\u)\in \R \times \R^3$, with $\alpha > 0 \wedge \|\u\| = 1$ by $\rotV = \alpha \u$
This is precisely the interpretation used in \cite{shuster93}.
After fixing this relationship the choice between \hqmcH and \hqmcS now defines different mappings from the rotation vector to the quaternion as expressed in \prettyref{tab:comparison}.

\paragraph{Angular velocity}
As a physical concept it is one of the more complex to introduce.
For the purpose of this paper it is enough to define it in coordinates through the following well-known kinematic equation, given some DCM-trajectory $\M(t)$ transforming coordinates with respect to frame $A$ into $B$-coordinates, with the coordinate frames as subscripts:
\begin{equation}
\label{eq:angularVel}
\crossMat{\avel_A} := -\M^{-1} \dot \M, \;
\crossMat{\avel_B} = -\dot \M \C^{-1}
\end{equation}
The second equation is already a consequence of the first, by means of the assumed transformation property of $\M$.
}

\subsubsection{Formal comparison independent of the usage}
\label{sec:comparison}
\def\Q{\mbf Q}
In \prettyref{tab:comparison} we compare the two homomorphic conventions\longversion{ based on important identities}.
The $\rotV$ denotes a \emph{rotation vector} in $\R^3$.
\longversion{%
The symbol pair ($\M, \star$) denotes either \hqmcH or \hqmcS depending on the column of the right hand side of the formula.
}%
\def\convF{\alpha_{\M}}%
\def\convRV{\alpha_{\rotV}}%
The \emph{convention-factors} $\convF,\convRV \in \{-1, 1\}$\footnote{PWTB \& DCM $\simeq -1$; active \& PBTW $\simeq 1$} encode the usage of rotation matrices ($\convF$) and rotation vectors ($\convRV$).
The usage of the quaternion follows from $\convF$ given the chosen $(\M, \star)$ and is identical to $\convF$ for homomorphic QM-conventions.
We'll look deeper into their effect in \prettyref{sec:big-picture-comparison}.
The indices $A$ and $B$ for the angular velocities, $\avel$, indicate differences stemming from various usage choices regarding $\avel$.
Depending on these choices one of the two formulas is correct.

\begin{table}[ht]
\caption{Comparing homomorphic QM-conventions}
\centering
\begin{tabular}{ r | c  c }
Expression & \multicolumn{2}{c}{Equal expression given QM-Convention} \\
& Hamilton, $\MH$ & Shuster, $\MS$  \\
\hline
$\M, \star$  & $\different{\MH}, \different{\odot}$ & $\different{\MS}, \different{\otimes}$  \\
\hline
$\!\M(\p) \M(\q)$  & $\different{\MH}(\p\different{\odot} \q)$ & $\different{\MS}(\p\different{\otimes} \q)$ \\
$\M(\q) \x $  & \multicolumn{2}{c}{$\imc{\q \star \imcinv{\x}\star  \q^{-1} }$}  \\
$\imc{\p \star \q} $  & $p_1\qv + q_1\pv \different{+} \pv \times \qv$& $p_1\qv + q_1\pv \different{-} \pv \times \qv$  \\
$\mbf i \star \mbf j $  & $ \different{+}\mbf k$ & $ \different{-}\mbf k$  \\
$\M(\q)$  & $q_1^2 \qid \different{+} 2q_1 \crossMat{\qv} + \Q(\q)\;\footnotemark{} $ & $ q_1^2 \qid \different{-} 2q_1 \crossMat{\qv} + \Q(\q)\;\footnotemark[\value{footnote}]{} $ \\
\hline
$\M(\rotV) $ & \multicolumn{2}{c}{ $\exp(+\convF\convRV\crossMat{\rotV}) $}\\
$\Rightarrow\footnotemark{}\q(\rotV) $  & $\pm\exp(\good{+}\convF\convRV\rotV / 2)$ & $\pm\exp(\highlight{-}\convF\convRV\rotV / 2) $  \\
\hline
\multirow{2}{*}{$\convF\dot\M$} & \multicolumn{2}{c}{$+\M \crossMat{\avel_A}$}\\
                           & \multicolumn{2}{c}{$+\crossMat{\avel_B} \M $}\\
\multirow{2}{*}{$\Rightarrow\footnotemark[\value{footnote}]{}\convF\dot\q$} & $\good{+} \frac 1 2 \q \different{\odot}\imcinv{\avel_A}$ & $\highlight{-}\frac 1 2 \q \different{\otimes} \imcinv{\avel_A}$\\
                           & $\good{+} \frac 1 2 \imcinv{\avel_B}\different{\odot}\q$ & $\highlight{-}\frac 1 2 \imcinv{\avel_B} \different{\otimes}\q $\\
\end{tabular}
\label{tab:comparison}
\end{table}
\addtocounter{footnote}{-1}
\footnotetext{$\Q(\q) := (\crossMat{\qv})^2 + \qv {\qv}^T $}
\addtocounter{footnote}{1}
\footnotetext{$\M(\rotV), \q(\rotV)$ and $\M(t), \q(t)$ are coupled here through $\M(\q)$.}

The neutral differences are highlighted in blue.
Green vs. red highlights the benefits of \hqmcH in terms of more similarity between corresponding rotation matrix and quaternion equations.
The table clearly shows that \hqmcH yields more formal similarity with the rotation matrices when related to rotation vectors or angular velocities:
The alternative, \hqmcS, does introduce a difference in sign for the expressions involving $\rotV$ or $\avel$ ($\highlight{-}$ instead of $\good{+}$) compared to the  matrix cases.
This difference in similarity with the corresponding rotation matrix formula does not depend on the interpretation/usage (encoded in $\convF$, $\convRV$, $A,B$) but only on the choice of the \compactversion{homomorphic QM-convention}\longversion{HQMC}\longref{ (see \prettyref{app:sim_proof})}.

\prettyref{tab:comparison} also indicates that additionally flipping the cross product (negating the $\crossMat{\cdot}$) would change the similarity argument in favour of \hqmcS.
In fact it seems probable that this lack of formal similarity was one reason to introduce the matrix valued operator $[[\rotV]]:= -\crossMat\rotV$ in \cite[p. 445]{shuster93}.
Unfortunately, this is hard to verify because it was introduced only because this \enquote{matrix turns out to be more convenient overall}.
In \cite{shuster2008} this matrix was abandoned again\longversion{ in favour of $[\rotV \times] := \rotV^\times$, which had already been mentioned in \cite[p. 445]{shuster93} with the words \enquote{Some authors define instead $[\rotV \times] := - [[\rotV]]$}}.

\subsubsection{Formal comparison distinguishing active and passive usage}
\label{sec:big-picture-comparison}
A further symmetry aspect in favor of Hamilton's multiplication becomes apparent if we compare both active and passive usage of matrices and quaternions (see \prettyref{sec:usage} and \prettyref{fig:rotation_conversion}).
This yields the \prettyref{tab:big-picture}, again using the factor $\convRV$ to encode the usage of the a given rotation vector $\rotV$.
In contrast to \prettyref{tab:comparison}, which assumes an arbitrary but single usage for rotation matrices and quaternions, it compares both usages (corresponding values in column 2 and 3).
Each row corresponds to a rotation representation: rotation matrix and three different flavours of rotation quaternions, each yielding homomorphic conversion to the rotation matrix in \prettyref{row:vector2matrix}.
The table shows that neither using Shuster's multiplication for the passive and Hamilton's for active usage (\prettyref{row:mixed_conversion}), nor a full switch to Shuster's multiplication (\prettyref{row:shuster_only}) can yield the same level of form-similarity, when compared to the rotation matrices, as the proposed Hamilton only (\prettyref{row:hamilton_only}) option (similarity flaws are marked in red vs. green).

\def\bigPictureRotV{\convRV\rotV}
\begin{table}[ht]
\newcounter{rowcount}
\setcounter{rowcount}{0}
\renewcommand{\arraystretch}{1.0}
\caption{
Active vs. passive usage given a rotation vector $\rotV$.%
}
\centering
\begin{tabular}{@{{\footnotesize\the\numexpr\value{rowcount}+1\relax})\hspace*{\tabcolsep}}>{\refstepcounter{rowcount}} r | c | c | c}
\multicolumn{1}{c|}{Rot. rep. }& Active & Passive & Composition \\
\hline
\label{row:vector2matrix}
$\M(\rotV)$ & $\exp(+\bigPictureRotV^\times)$ & $\exp(-\bigPictureRotV^\times)$ & $\cdot$ \\
\hline
\multicolumn{4}{c}{ Homomorphic rotation quaternion options:}\\
\hline
\longversionStrict{\multicolumn{4}{c}{ Hamilton only (proposed)}\\}
\label{row:hamilton_only} $\q(\rotV)$& $\exp(\good{+}\bigPictureRotV / 2) $ & $\exp(\good{-}\bigPictureRotV / 2)$ & $\good{\odot}$ \\
\hline
\longversionStrict{\multicolumn{4}{c}{ Hamilton (active) \& Shuster (passive)}\\}
\label{row:mixed_conversion}$\q(\rotV)$ & $ \exp(\good{+}\bigPictureRotV / 2) $ & $ \exp(\highlight{+}\bigPictureRotV / 2) $ & \color{red}{$\odot$(act), $\otimes$(pass)}\\
\hline
\longversionStrict{\multicolumn{4}{c}{Shuster only} \\}
\label{row:shuster_only} $\q(\rotV)$ & $\exp(\highlight{-}\bigPictureRotV / 2) $ & $\exp(\highlight{+}\bigPictureRotV / 2)$ &  $\good{\otimes}$\\
\hline
\end{tabular}
\label{tab:big-picture}
\end{table}

\longversion{
\discuss{Should the following go to the history appendix?}
We could not find a single reference from back then that distinguishes active and passive rotation quaternions despite the fact that this distinction was well-known at the time for rotation matrices.
This seems odd today as rotation quaternions are applied to coordinates fully analogous to rotation matrices.
Historically this could stem from the fact that quaternions were apparently misunderstood as consisting of ``physical'' vectors and scalars \cite{shuster93, shuster2008}.
Given this misunderstanding the two usages made no sense.
One of the important contributions of \cite{shuster93} was in fact to free rotation quaternions from this misconception for the aerospace attitude estimation and control community \cite{shuster2008}.
Possibly this insight came to late to prevent the problematic abuse of active quaternions for the passive usage.

Another important reason could have been that the rotation vector to quaternion conversion was initially copied while overlooking that it was using active quaternions.
This is strongly indicated by how \cite{shuster93} writes about \cite{battin1987} that clearly uses active quaternions (see also \prettyref{app:history}).
}

\longversion{
\subsubsection{Why care about formal similarity between matrices and quaternions?}
There is mostly one problem with little flaws in form similarities: they are error prone, as already found in \cite{shuster2008} to justify the change proposed in \cite{shuster93}:
\quotebig{%
System development will be less prone to error if the multiplication rule for quaternions has the same order as that for the corresponding rotation matrices.}

This might seam like a weak argument and to many it probably seems that it isn't enough to change an old convention for.
But it was a lack of similarity that started the whole discussion in the first place.
One could have just accepted that quaternions were multiplied ``the other way'', \eqref{eq:problem}.
The first change of convention to \hqmcS was motivated and solely justified by a gain in similarity (same multiplication order).
Switching to \hqmcH instead would have yielded the same gain concerning multiplication order plus the additional similarity in sign highlighted in \prettyref{tab:comparison}.

However, the major and most important difference remains in with which convention the two alternatives break:
The former with the quaternion multiplication order and the latter with the correspondence between rotation quaternions and rotation matrices.
These conventions are very different. 
Most importantly in how close they are to application vs. pure math.
In general it is much better to first question the convention closer to application because typically these convention have not settled over such a long time and it did not necessarily had to prove itself withing a bigger picture.
Another important difference back then was that the quaternion multiplication was undoubted while the quaternion to matrix conversion was already defined in competing ways as M. D. Shuster observed himself in \cite[p. 473]{shuster93} (see also \prettyref{app:history}).

\subsection{Algorithmic and numeric perspective}
\label{sec:perspective_algorithmic}
From an algorithmic perspective sometimes even representations that are fully isomorphic are still not equally preferable for example because one might better fit the machine architecture.
However, this particular choice between \hqmcH and \hqmcS is very unlikely to have any impact on numeric efficiency or stability.
This can be seen as follows:
Any existing algorithm using one of the two HQMCs can be transformed into an algorithm using the other convention by applying $\Phi$ from \eqref{eq:antiisomorphism} to all relevant quantities (see \prettyref{sec:migration_translation}).
It is apparent that the effective numerical difference between versions of the algorithms is a sign flip on either the real or imaginary part for all quaternions involved.

The coordinate-vectors to be rotated do not need to be changed at all as shown in \prettyref{mex:quatrot}.
In summary the only difference remaining is the partial sign flips on the quaternions.

While tedious to proof formally, it seems intuitively convincing for this change not to be able to introduce numerical instability because it is only a partial mirroring of a parameter space.
Concerning run-time efficiency it should not make any significant difference because almost for any usual operation the sign flip can be statically merged into other operations, such as scaling or addition without any extra cost.

\subsection{Intuitive perspective}
\label{sec:perspective_intuitive}
One very special purpose of rotation representation is to allow an intuitive understanding of the space of spatial rotations.
The axiomatic requirements, to preserve the origin, lengths and angles, is good to understand what a rotation does.
But in order to imagine the full range of rotations parametrizations can be very useful tools.
In this section, we will refer to this use case of a parametrizations as the ``rotation range intuition" (=:RRI).
The most useful for RRI is probably the angle axis concept and the closely related rotation vector since they immediately yield the very intuitive concepts of a rotation axis and a rotation magnitude and a rotation direction.
One could see unit quaternions as almost as useful for that task because they are quite close to an angle axis: 
their imaginary part does indeed indicate the axis of rotation and both real and imaginary part the magnitude of rotation.
And the direction can be retrieved similarly to angle axis.
However, at least in order to retrieve the actual magnitude an application of the arc cosine is required.

For those who really use quaternions as an intuitive device the decision between \hqmcH and \hqmcS has relevance for RRI because the rotation direction flips when inverting the quaternion.
This might be a historical reason for why the relation $\q = \exp(-\rotV / 2)$ seemed wrong and was rejected as an alternative.
Here a two arguments that counter this argument:

\paragraph{Use pain angle-axis instead}
Since computers can easily translate any quaternion convention into an angle-axis pair there is no need for any human to work with quaternions intuitively.
We propose to solely lean on angle-axis or rotation vectors for RRI and similar intuitive tasks and use algorithms to convert between them and other representations wherever necessary.

\paragraph{Consider the active usage}
Many scientists, for example from other fields such as mathematics or physics, are used to think intuitively of rotations from the active perspective (see \prettyref{app:rotation_usage}).
For those it is natural to use quaternions actively, i.e. to rotate vectors by the means of their coordinates instead of translating coordinates between coordinate frames.
Such an active quaternion must be inverted when compared to the passive quaternion often used in robotics or attitude estimation and control, assuming that everybody agrees on one multiplication.
Therefore both equations, $\q_1 = \exp(-\rotV / 2)$ and $\q_2 = \exp(\rotV / 2)$ will be inevitably used somewhere.
The decision between \hqmcH and \hqmcS can only decide which of the two perspectives has to live with the minus.
Of course, one could argue here that this is the very reason to have two multiplications and only one formula, $\q = \exp(\rotV / 2)$.
But the fact that there are active and passive perspectives on rotations is unavoidable and useful as it allows a truly different perspective.
Why should it not be reflected in conversion formulas? 
It is reflected in the corresponding to matrix formulas (see \prettyref{sec:big-picture-comparison}).
In fact the minus in $\q_1 = \exp(-\rotV / 2)$ can be interpreted as reminiscence to the transitions from an active, $\rotV$ to a passive $\q_1$, given the \hqmcH choice.
The other formula, $\q_2 = \exp(\rotV / 2)$, would in that case describe the relation between an active quaternion, $\q_2$, and an active $\rotV$.
On the other side two multiplication are neither useful nor unavoidable both because they merely yield an (anti)isomorphic structure.
} 

\section{Recipes}\label{sec:recipes}
In this section we give some practical advice to a) identify QM-conventions and b) migrate between them.

\subsection{How to detect which QM-convention is used}
\longversion{Given some convention, $C$, it can be important to find out which QM-convention is actually included, i.e. which minimal QM-convention $D$ is included in $C$ ($D\subseteq C$).}
\longversion{%
This could be for example in order to find out whether it is necessary to translate formulas (\prettyref{sec:migration}) or conjugate quaternion values to interface with a piece of software.
As already mentioned in \prettyref{sec:natural-convention-order} it is sufficient to check whether the axioms of a candidate convention, $D$ are compatible with $C$ to conclude that $D\subseteq C$.
}%
\longversion{All interesting minimal }QM-conventions are determined by two binary choices, which quaternion to matrix map, $\M$, and which quaternion multiplication, $\star$.
Therefore, it is enough to determine these two choices, which can be done as follows\longversion{ (see \prettyref{app:detection-formulas} for further details)}:
\subsubsection{For the \textbf{quaternion multiplication}} one of the easiest ways is to find out the result of the product $\mbf i \star \mbf j$.
If the result is $\mbf k$ then $\star = \odot$ otherwise it is $-\mbf k$ and $\star = \otimes$%
\longversion{\footnote{
This assumes one of the two multiplications are used.
Otherwise the $C$-``quaternions'' either fail to be a four dimensional division algebra over $\R$ and should not be called quaternions, or the $\mbf i, \mbf j, \mbf k$ are wrongly assigned (this follows immediately from the Frobenius theorem).
} (see also \prettyref{tab:comparison})}%
.

\subsubsection{The \textbf{quaternion to matrix conversion}} can be identified by applying it to a suitable test quaternion, e.g. $\q_T:=\sqrt {0.5} (1, 0, 0, 1)^T$.
It holds $\MH(\q_T) = \M_T := 
\left(\begin{smallmatrix}
0 &-1 &0 \\
1 &0 &0 \\
0 &0 &1
\end{smallmatrix}\right)$ and $\MS(\q_T) = \M_T^T$\longversion{ (see \prettyref{app:eulerRodrigues})}.
This test can also be performed on a \textbf{matrix to quaternion conversion}.
It maps $\M_T$ to $\pm \q_T$ if $\MH$ is used and to $\pm \overline {\q_T}$ in case of $\MS$%
\longversion{\footnote{Otherwise either the $C$-quaternions are no quaternions or the rotation given a quaternion is not defined with one of the two usual possibilities \eqref{eq:defQuatMatMaps}.}}%
.

\subsection{Migrating from one QM-convention to another}\label{sec:migration}

There are several efficient ways to \emph{migrate} a \emph{tool} (e.g. formulas, \longversion{derivations, }proofs, publications or algorithms and implementations) between QM-conventions, where migration shall refer to a transformation after which the tool does exactly the same job but with all quaternions going in or out being compatible with the target convention.
We introduce here two powerful alternative procedures, \emph{translate} and \emph{interface}, that are applicable to all types of tools.
Both procedures migrate between the two homomorphic QM-conventions%
\footnote{To migrate a non homomorphic QM-convention first replace all multiplications with the other and flip their arguments. This is an equivalence transformation yielding a tool using a homomorphic convention.}%
.
It can be very helpful to first decompose a tool into smaller \emph{sub-tools} and migrate each with the most advantageous procedure.
\longversion{
For this it is crucial that the boundaries of the components composing the tool are clear and yield a distinct and total decomposition.
For an equation any directed acyclic graph (DAG) graph of logic, predicate and function evaluation expressions, including \emph{nullary} expressions (i.e. constants or variables), resembling the same equation, would be a candidate decomposition, no matter how big the individual expressions are (highly ambiguous!).
An algorithm is usually already decomposed into functions, but it might help considering a different decomposition.
}

\subsubsection{The two migration procedures}
See \prettyref{app:migration_proof} for a sketch of a correctness proof.
For both procedures a successive mathematical simplification step\longversion{ (mostly merging in quaternion conjugations)} is recommended\longversion{ to avoid complexity inflation}.

\paragraph{Translate}\label{sec:migration_translation}
Replace all quaternion\longversion{ valued} \textbf{constants}, $\mbf {c\in \H}$, within the tool with their \textbf{conjugated} value, $\bar{\mbf c}$, and all quat. \textbf{multiplications} with their \textbf{flipped} version ($\otimes\!\leftrightarrow\!\odot$). 

\paragraph{Interface}\label{sec:migration_interfacing}
\textbf{Conjugate} all \textbf{quaternion} valued \textbf{in-} and \textbf{outputs} of the tool\footnote{Also suggested in \cite{naif2003}.}.
Components of a \emph{partial} quaternion (i.e. not constituting a complete quaternion) going in or out should be treated as plain real numbers\longversion{ and not changed}.
\longversion{
For these components it is important to treat them consistently: it must not happen that e.g. a real number going out of one component is considered part of a full quaternion on 
a receiving component. 
If this seems to happen an additional sub tools needs to be defined doing the assembly from real number (components / coordinates) into a quaternion and that sub tools must be migrated as well.
Please note that interfacing does not change the interior of the tool.
While this can save effort it does not get rid of the other convention, it is just contained within the tool.
For most tools this is therefore more a temporary solution%
\footnote{For formula like tools and in combination with a mathematical simplification step it can simplify a successive translation step.
In such cases it is an interesting option even for a complete migration.}%
.
}

\subsubsection{Examples}
Next, we provide some examples\longversion{, mostly taken from \prettyref{tab:comparison}, while suggesting efficient ways to migrate them}:

\paragraph{$\mbf i\otimes \mbf j = -\mbf k\,$}
translates into $-\mbf i\odot -\mbf j = -(-\mbf k)$, which simplifies to $\mbf i\odot \mbf j = \mbf k$.
Interfacing it would not change it since it has no inputs and only a logic output\longversion{, which should and would remain true}.

\paragraph{$\imc{\q}$}\label{mex:imc}
interfaces into $\imc{\bar \q} = - \imc{\q}$.
When translating it, its implicit dependence on constants must be respected. 
One way to define its returned triple $(\alpha)_{i=2}^4 \in \R$ is as part of the unique solution to $\q = \sum_{i=1}^4 \alpha_i {\mbf e}_i$, with $\alpha_1 \in \R$. 
The equation translates into $\q = \sum_{i=1}^4 \alpha_i \overline{{\mbf e}_i}$, which has the solution $\bar {\mbs \alpha}$. 
Hence, translation also yields $-\imc{\q}$.

\paragraph{$\dot \q = -\frac 1 2 \q \otimes \imcinv{\avel}$}
could be first decomposed in $\dot \q = -\frac 1 2 \q \otimes \,\cdot$ and $\imcinv{\avel}$.
The first translates into $\dot \q = -\frac 1 2 \q {\odot} \,\cdot\,$\footnote{The variable $\q$ is an input and no constant.}.
\longversion{
Interfacing it would instead yield: $\dot {\bar \q} = \frac 1 2 \bar \q {\otimes} \,\bar \cdot \Leftrightarrow \dot {\q} = -\frac 1 2   \cdot\,{\otimes} \q$, where the right side is different but equivalent to $\frac 1 2 \q {\odot}\,\cdot\,$\footnote{The result of a migration is not the same for all ways, but must be equivalent according to our definition (does the same job)}.
}%
The second interfaces into $\overline{\imcinv{\avel}} = -\imcinv{\avel}$\longversion{ or translated into $\omega_1 (-\mbf i) + \omega_2 (-\mbf j) + \omega_3 (-\mbf k) = -\imcinv{\avel}$}.
Putting together yields the expected $\dot \q = \frac 1 2 \q \odot \imcinv{\avel}$.

\paragraph{$\MS(\q)$}\label{mex:quatmat}
Interfaces into $\MS(\bar \q) = \MH(\q)$\footnote{Hence, the matrix part of the QM-convention is indeed migrated.}.
It translates into the same, because $\MS$ must use coordinates with respect to a basis $(\q_i)_1^4$ (constants) for its input and $\sum \alpha_i \bar{\q_i} = \q \Leftrightarrow \sum \alpha_i \q_i = \bar \q$, because $\alpha_i \in \R$ (compare $\ref{mex:imc}$).

\paragraph{$\imc{\q \otimes \imcinv{\x}\otimes \q^{-1}}$}\label{mex:quatrot}
could be first decomposed in $\imc{\cdot}$, interfacing into $\imc{\bar\cdot} = -\imc{\cdot}$, and $\q \otimes \imcinv{\x}\otimes \q^{-1}$,
translating into $\q \odot (-\footnote{From translating constants in $\imcinv{\x} = x_1 \mbf i + x_2\mbf j + x_3 \mbf k$}\imcinv{\x})\odot \q^{-1}$.
Putting together yields the expected result, $\imc{\q \odot \imcinv{\x}\odot \q^{-1}}$.

\section{Conclusion}
In this paper, we suggested an alternative solution to the antihomomorphy problem that lead to the introduction of the flipped quaternion multiplication.
We further proposed to discontinue its use in favour of Hamilton's original definition combined with the suggested alternative solution.
To argue for this we gave evidence of the cost of maintaining two multiplications and showed that for principal reasons there cannot be any significant capability benefit in using the flipped multiplication for theory or algorithms.
\longversion{%
Only those few who might (still) use quaternions to support their rotation intuition and use the passive world to body perspective have objective reason to prefer the flipped multiplication.
To those we can only recommend to use angle axis parameters since they are even better suited for intuition.
}%
Additionally, we demonstrated that the formal similarity between matrices and quaternions, when related to angular velocity or rotation vectors, is in favour of the Hamiltonian multiplication independently of any other conventional decision as long as the definition of the cross product is not additionally questioned.
Furthermore, we provided recipes for how to migrate formulas and algorithms from one quaternion multiplication to the other as well as how to detect which convention is used in a given context.

\section*{Acknowledgment}
The authors are grateful to Michael Burri and Anne-Katrin Schlegel for very helpful comments.

\appendix
\renewcommand{\thesubsection}{\Alph{subsection}}
\setcounter{subsection}{0}
\renewcommand*{\theHsection}{appendix.\the\value{section}}
\subsection{Proof sketch: Correctness of migration recipe}
\label{app:migration_proof}
A full technical proof is far beyond the scope of this short paper and we only provide a sketch of a proof:
To show the claim we assume the opposite and that we have a \emph{counter example}.
I.e. a tool, $T$, that after some migration, $M$, as in \prettyref{sec:migration} into $M(T)$ is not doing the same job when used with quaternions of the target convention.
\newcommand\fail[2]{{{#1}\not\simeq{#2({#1})}}}
Formally, $M(T)$ does not yield the corresponding, i.e. quaternions in the target convention and otherwise equal, outputs when given corresponding inputs (denoted with $\fail{T}{M}$).

\newcommand\doesMigrateAsAWhole[2]{{{#1}\;\mathtt{doesMigAsAWhole}\;{#2}}}
\newcommand\doesTranslate[2]{{#1}\;\mathtt{doesTranslate}\;{#2}}
\def\hasQuaternionIO{{\mathcal{Q}}}
\subsubsection{T can be assumed migrated as a whole by M}($\doesMigrateAsAWhole{M}{T}$)
If not we take the first sub tool, $T'$, on the way from input to output, for which $\fail{T'}{M}$, as new $T$ until $\doesMigrateAsAWhole{M}{T}$.
\subsubsection{T can be additionally assumed having no quaternion in- or outputs and being translated by M} 
($\neg\hasQuaternionIO(T) \wedge \doesTranslate{M}{T}$)
To ensure that we inflate $T$ into $\hat T$ by adding bijective conversions per quaternion in- or output, $\q$, into a pair $\in \R \times \R^{3\times 3}$ through $\M$\footnote{E.g. $(\sign(\Re(\q)) \|\q\|, \M(\q  \|\q\|^{-1}))$ if $\q \not = 0$, otherwise $(0, \mbs 0)$} (from the source convention).
$\hat T$ can then be used for exactly the same jobs by employing inverse conversions wherever quaternions were exchanged with $T$.
We migrate $\hat T$ by $\hat M$ into $\hat M(\hat T)$ by migrating the contained $T$ using $M$ and translating the inflation layer (effectively translating all source $\M$ into the target $\M$; see \prettyref{mex:quatmat}).
Since $\hat M(\hat T)$ would be used through conversions using the target $\M$, which would cancel out the translated inflation, causing the same effective input output behaviour as $M(T)$ and therefore different from $\hat T$. 
It follows $\fail{\hat T}{\hat M}$, while $\neg \hasQuaternionIO(\hat T)$.
Furthermore $T$ must have been already translated:
If it had been interfaced (as a whole) $\hat M(\hat T)$ would exactly behave as $\hat T$, because the interfacing conjugation effectively revokes the translation of the inflation, rendering $\hat M$ into an equivalence transformation, contradicting $\fail{\hat T}{\hat M}$.
It follows $\doesTranslate{\hat M}{\hat T}$ and, since component-wise translating is equivalent to translating as a whole also $\doesMigrateAsAWhole{\hat M}{\hat T}$.

\subsubsection{T is no counter example}
It is well-known and straight forward to verify that conjugation is a structure isomorphism $(\H, \plus, \odot, \bar \cdot) \simeq (\H, \plus, \otimes, \bar \cdot)$ that lets the embedding $\R \subset \H$ invariant.
Since $T$ can only use the quaternions by means of their structure and their relation to $\R$ (nothing more is defined about them after all) replacing all quaternion constants according to the isomorphism and all operations with their isomorphic partners, as precisely done by the translation procedure, cannot change its input-output behavior for non-quaternions.
Therefore $\neg\hasQuaternionIO(T) \wedge \doesTranslate{M}{T}$ contradicts $\fail{T}{M}$ and proves the claim.

\longversion{%

\subsection{Proof: Hamilton QM-convention always yields more similarity}\label{app:sim_proof}%
Let $C$ be a consistent partial convention that is a homomorphic QM-convention and additionally determines a mapping, $C.\q : \R^3\to\U$ from rotation vectors to the unit quaternions.

\subsubsection{Rotation vector to matrix and rotation quaternion conversions}
\def\v{\mbf{v}}
As $C$ is consistent it holds for all $\rotV\in \R^3$ that $C.\M(C.\q(\rotV)) = C.\M(\rotV)$, with $C.\M$ denoting the rotation vector to matrix conversion determined by $C$.
  Assuming the right side has the form 
  \begin{equation}
  \label{eq:sim_proof_angle}
  C.\M(\rotV) = \exp (\mbf a(\rotV)^\times),
  \end{equation}
  with some $\mbf a : \R^3 \to \R^3$ \footnote{We are not aware of any other existing convention. Also, otherwise it should be hard to justify $\rotV$ as rotation vector unless the rotation matrix does not act through the matrix product.}
  it holds necessarily 
  \begin{equation}\label{eq:sim_proof_m1_consistency}
  \begin{aligned}
  \MH(\exp(\imcinv {\mbf a(\rotV) / 2})) 
    &= \exp (\mbf a(\rotV)^\times) = C.\M(\rotV) \\
    &= C.\M(C.\q(\rotV))
  \end{aligned}
  \end{equation}
  using the exponential map of quaternions, $\exp$ \footnote{The exponential map is the same for both, Hamilton's and Shuster's multiplication.},
  because for all $\x\in \R^3$ it holds, independently of $C$, that 
  \[\MH(\exp(\imcinv{\x / 2})) = \exp(\x^\times). \]
  
  Due to the definition of $\MH$ it follows, for the case that \hqmcH is sub convention of $C$, (and therefor $C.\M|_{\U} = \MH$):
  \begin{equation}
  \label{eq:sim_proof_case1}
  C.\q(\rotV) = \alpha \exp(\imcinv{\mbf a(\rotV) / 2}),
  \end{equation}
  with $\alpha \in\{-1, 1\}$.
  To get the analogous result for the second case, \hqmcS is sub convention of $C$, we can translate (see \prettyref{sec:migration}) the left hand side of \eqref{eq:sim_proof_m1_consistency} into the other QM-convention and get
  \begin{equation}
  \label{eq:sim_proof_m2_consistency}
  \MS(\exp(\highlight{-}\imcinv {\mbf a(\rotV) / 2})) = C.\M(C.\q(\rotV)),
  \end{equation}
  which yields analogously for this case
  \begin{equation}
  \label{eq:sim_proof_case2}
  C.\q(\rotV) = \alpha \exp(\highlight{-}\imcinv{\mbf a(\rotV) / 2}).
  \end{equation}
  
  Comparing \eqref{eq:sim_proof_case1} and \eqref{eq:sim_proof_case2} with \eqref{eq:sim_proof_angle} it becomes evident that the second case, \hqmcS being part of $C$, always yields an extra $-$.

\subsubsection{Angular velocity kinematic equation}
  Our definition for the angular velocity, \eqref{eq:angularVel}, was for a specific interpretation of the rotation matrix $\M$ and assignment of frames.
  However, all common conventions and frame assignments yield very similar equations. In fact to the best of our knowledge they all have one of the two forms:
\begin{equation} 
\avel^\times = \beta\M^{-1} \dot \M, \; \text{or} \;
\avel^\times = \beta\dot \M \M^{-1},
\end{equation}
with $\beta\in \{-1, 1\}$ and $\M(t)$ a rotation matrix trajectory.

For the following we assume the first form. The claim can be shown for the second analogously.

Firs we observe that a QM-convention, $\q\mapsto \M(\q)$, completely determines the corresponding relation to the quaternion trajectory, $\q(t)$ representing the same trajectory as $\M$:
\begin{equation}
\label{eq:omegaCQ}
\avel^\times = \beta\M^{-1} \dot \M = \beta \M^{-1}(\q)\dot{\M(\q)} = \gamma \frac12\beta \q^{-1}\dot\q
\end{equation}
with a convention-factor $\gamma := 1$ for \hqmcH and $\gamma := -1$ for \hqmcS.

The last equality of \eqref{eq:omegaCQ} is not trivial but can be extracted from known identities as follows.
\prettyref{tab:comparison} contains a well-known special case of \eqref{eq:omegaCQ} for $\beta = -1$ and $\avel = \avel_A$:
\[-\M^{-1}(\q) \dot {\M(\q)} = \avel_A^\times = -\frac12\gamma\q^{-1}\dot\q\]

Multiplication with $-\beta$ proofs the last equality of \eqref{eq:omegaCQ}.
And comparing \eqref{eq:omegaCQ} with the assumed $\avel^\times = \beta\M^{-1} \dot \M$, shows that $\gamma=1$ yields more similarity independently of $\beta$ and therefore any conventional decision beyond the QM-convention.

\subsection{Representing 3d proper rotations with matrices}\label{app:rotation_matrices}

In this section we aim at clarifying the different kinds of ``rotation matrices'' and the distinction about active and passive rotations as far as needed for the arguments presented in the rest of the paper. 
For that we first we briefly define the setup and some notions we need in order to define rotation matrices.
Second we define rotation matrices in three different ways.
After that we explain their relation to the usages, active and passive, and how those are related to the composition of rotations.

\def\V{\text{E}_3}
\subsubsection{Rotations in 3d-Euclidean space}
First we need a general setup in which to define rotations in three dimensions.
The Euclidean three dimensional vector space $\V$ with the scalar-product $\langle \cdot , \cdot \rangle$ is sufficient.
A linear map $\Phi : \V \rightarrow\V$ is a proper%
\footnote{We will often skip the ``proper'' for brevity. We are not interested here in improper rotations.} 
rotation iff it preserves the scalar product and the orientation. 
It is important to note, that $\Phi$ is not a matrix, but just a map mapping abstract vectors to abstract vectors.
A basis, $\mathcal B = (\mbs b_i)_{i=1}^3$, we call a \emph{positively oriented orthonormal basis (PONB)} iff it is of positive orientation and orthonormal, i.e.
\[ \langle \mbs b_i, \mbs b_j \rangle = \delta_{ij}\]

And we write $[\mbs v]^\mathcal B\in \R^3$ to denote the coordinates of a vector $\mbs v\in\V$ with respect to $\mathcal B$, such that $\mbs v = \sum_{i=1}^3 [\mbs v]_i^\mathcal B \mbs b_i$ or equivalently, because $\mathcal B$ is orthonormal, 
\[[\mbs v]_i^\mathcal B = \langle \mbs v, \mbs b_i \rangle\]

\subsubsection{Three competing major approaches to represent a rotation with a matrix}
\label{app:matrix-representation}
Given a PONB, $\mathcal B = (\mbs b_i)_{i=1}^3$ there are three major approaches of representing a rotation $\Phi$ with a matrix:

\begin{enumerate}[(a)]
\item The \emph{\underline representing matrix}\footnote{Yes, this name seems a bit unfair - compared to the others, since we are talking about how to represent a rotation with a matrix. But the other common name for that matrix connected to a general linear map is ``transformation matrix''. 
Unfortunately this name is often specialized to the matrix representing elements of the special euclidean group --- at least in the robotics community.} of the rotation with respect to a PONB $\mathcal B$, such that for all vectors $\mbs v\in \V$ :
	\begin{equation}
	\label{eq:representing-matrix}
		[\Phi(\mbs v)]^\mathcal B = \mbf R_\Phi^\mathcal B [\mbs v]^\mathcal B
	\end{equation}
	This is typically used in mathematical literature. But one could have flipped $\mbf R_\Phi^\mathcal B$ and $[\mbs v]^\mathcal B$ interpreting the latter as row-vector
	\footnote{The usual identification of $\R^3$ with $\R^{3\times 1}$ is purely conventional.}
	and define this way the transposed matrix.
\item The \emph{change-of-\underline basis matrix} that transforms the coordinates of all fixed vectors $\mbs v\in\V$ when the PONB $\mathcal B$ is rotated, with respect to which the coordinates of $\mbs v$ are given before and after the rotation:
	\[ [\mbs v]^{\Phi(\mathcal B)} = \mbf B_\Phi^\mathcal B [\mbs v]^\mathcal B\]
	One could have exchanged $[\mbs v]^{\Phi(\mathcal B)}$ and $[\mbs v]^\mathcal B$, ending up with the inverse matrix ${\mbf B_\Phi^\mathcal B}^{-1}$, as used e.g. in \cite[p. 25]{Murray9400}.
\item The \emph{direction cosine matrix} that contains the coordinates of a rotated PONB $\mathcal B$ with respect to itself before rotation - one rotated basis vector per row:
	\[ (\C_\Phi^\mathcal B)_{i\cdot} = [\Phi(\mbs b_i)]^\mathcal B \text{ or } {\C_\Phi^\mathcal B}_{ij} := \langle \Phi(\mbs b_i), \mbs b_j \rangle\]
	Here $\mbf M_{i\cdot}$ denotes the $i$th row of a matrix $\mbf M$.\\
	This is used e.g. in \cite{Hughes8600}, p. 8 and  \cite{shuster93}, p. 447.\\
	One could have chosen columns instead of rows.
	This would have yielded the transposed matrix ${\C_\Phi^\mathcal B}^T$.
\end{enumerate}
In all three cases the resulting matrix is itself orthogonal (so its transposed is its inverse) and it has determinant 1 (this corresponds to $\Phi$ preserving the orientation). 
They all depend on the choice of the basis $\mathcal B$ but not one to one.
In general given a second PONB $\mathcal C$ which results from applying the rotation $\Psi$ on $\mathcal B$ the representing matrices are all connected by the following conjugation, while $\mbf M$ denotes one of $\mbf R, \mbf B, \C$:
\begin{equation}\label{eq:BaseChange}
\mbf M_\Phi^\mathcal C = \mbf B_\Psi^\mathcal B\mbf M_\Phi^\mathcal B {\mbf B_\Psi^\mathcal B}^{-1}
\end{equation}
But they are not all the same in value (in definition they are clearly different). In fact the following always holds:
\begin{equation}\label{eq:ABCRelation}
{\mbf R_\Phi^\mathcal B}^{-1} = \mbf B_\Phi^\mathcal B = \C_\Phi^\mathcal B
\end{equation}
Hence, considering only the values of $\mbf R, \mbf B, \mbf C$ leads to two different rotation matrices (for a given basis and a given rotation) always being a mutually inverse pair.

\subsubsection{Usage and Composition}\label{app:rotation_usage}

The value of $\mbf R$ is called the \emph{active} rotation matrix and the value of $\mbf B, \mbf C$ the \emph{passive} rotation matrix for the rotation $\Phi$ with respect to $\mathcal B$.
The motivation behind active is the fact that the definition of $\mbf R$ is based on how $\Phi$ rotates $\mathcal B$
while the idea behind passive is that $\mbf B$ is based on how the coordinates of a fixed $\mbf v$ change when switching basis.
We call these different ways to use a rotation matrix its \emph{usage}.
However, as we mentioned before, all the definitions include an arbitrary binary choice that taken differently would lead to the transposed / inverse matrix.
Therefore the assignment of active and passive to the values of rotation matrices seems arbitrary or purely conventional, too.
Especially it is unclear why the definitions are not aligned such that at least they all lead to the same matrix.
The reason is that for various applications the important difference between the available matrices is not their definition but what the standard matrix product corresponds to in terms of rotations.
Because for every pair of invertible matrices, $\mbf X, \mbf Y$ it holds $(\mbf X\mbf Y)^{-1} = \mbf Y^{-1}\mbf X^{-1}$ it actually makes a difference whether one uses the orthogonal matrix $\mbf R$ or its inverse, e.g. $\mbf C$.
And switching between the two options has the same effect on the result as flipping the two represented rotations or the two matrices.
Both resulting matrices correspond to a composition of two rotations, $\Phi, \Psi$ with a significant difference.
Using $\mbf R$ as active and $\mbf C$ as the passive matrix:

\begin{eqnarray}
	\mbf R_{\Phi\circ\Psi}^\mathcal B &\underset{\eqref{eq:representing-matrix}}=& \mbf R_\Phi^{\mathcal B} \mbf R_\Psi^\mathcal {B} \label{eq:ASingleBasis} \\ 
	&=& ({\mbf R_\Psi^{\mathcal B}} {\mbf R_\Psi^{\mathcal B}}^{-1}) \mbf R_\Phi^{\mathcal B} \mbf R_\Psi^\mathcal B \nonumber\\
	&=& {\mbf R_\Psi^{\mathcal B}} ({\mbf R_\Psi^{\mathcal B}}^{-1} \mbf R_\Phi^{\mathcal B} \mbf R_\Psi^\mathcal B) \nonumber\\ 
	&\underset{\eqref{eq:ABCRelation}}=& {\mbf R_\Psi^{\mathcal B}} (\mbf B_\Psi^\mathcal B\mbf R_\Phi^{\mathcal B} {\mbf B_\Psi^{\mathcal B}}^{-1}) \nonumber\\ 
	&\underset{\eqref{eq:BaseChange}}=& \mbf R_\Psi^\mathcal {B} \mbf R_\Phi^{\Psi(\mathcal B)} \label{eq:AMultiBasis}\\
	\C_{\Phi\circ\Psi}^\mathcal B 
	&\underset{\eqref{eq:ASingleBasis}, \eqref{eq:ABCRelation}}=& \C_\Psi^{\mathcal B} \C_\Phi^\mathcal {B}  \nonumber  \\ 
	&\underset{\eqref{eq:AMultiBasis}, \eqref{eq:ABCRelation}}=& \C_\Phi^{\Psi(\mathcal B)} \C_\Psi^\mathcal {B} \label{eq:CMultiBasis}
\end{eqnarray}

The active matrix, $\mbf R$, matches the composition order, $\Phi$ after $\Psi$, with its multiplication order when one keeps the basis of representation \eqref{eq:ASingleBasis}.
The inverse, passive matrix, $\mbf C$ does the same iff one represents the second rotation with respect to the second (rotated first) basis \eqref{eq:CMultiBasis}.
In other words the passive is better suited when from the applications domain it is preferable to switch the reference basis with each rotation (e.g. attitude estimation and control) and the active is preferable whenever it is preferable to keep the reference frame fixed for all the composed rotations.
In contrast to the definitions above this difference is not arbitrary or conventional.
And therefore the labels active and passive, i.e. the usage of a rotation matrix, would be more robustly defined using these properties with respect to composition of rotations.

The big confusion concerning rotation matrices is caused --- in our opinion --- by the fact that papers and textbooks from different disciplines seemingly fight for the authority to define which definition deserves the label ``rotation matrix'' instead of mentioning the two fundamental options and explicitly choosing the one that is more suitable for a given use case.
Of course somebody who only knows about one definition and usage of rotation matrices will be confused when reading texts that call the inverted matrix ``the rotation matrix''.

It is one plausible but hard to verify theory that this habit of ignorance ultimately lead to the mistake to use active rotation quaternions in attitude estimation and control (see \prettyref{sec:big-picture-comparison}) --- a domain where passive rotation matrices were and still are predominant used.
This mistake inevitably lead to the problem \eqref{eq:problem}, because --- as we have seen --- the essential difference between the two usages is precisely how they relate to the compositions of rotations.

\subsection{Formulas based on which QM-conventions can be detected}
\label{app:detection-formulas}
\subsubsection{Cross products to skew symmetric matrix}
\[
\mbf a^\times = \begin{bmatrix}\,0&\!-a_3&\,\,a_2\\ \,\,a_3&0&\!-a_1\\-a_2&\,\,a_1&\,0\end{bmatrix}, [[\mbf a]] =  \begin{bmatrix}\,0&\,\,a_3&\!-a_2\\ -a_3&0&\,\,a_1\\\,\,a_2&\!-a_1&\,0\end{bmatrix}
\]

\subsubsection{Quaternion to Matrix conversion}\label{app:eulerRodrigues}
Here we derive the active Hamilton-quaternion to active matrix or passive Hamilton-quaternion(world to body) to passive matrix mapping, which is equivalent to the passive Shuster-quaternion to active matrix and active Shuster-quaternion to passive matrix:
\begin{align}
\imquatvec{\q \x \q^{-1}} &= \imquatvec{(\q \x)\q^{-1}} \\
&= \imquatvec{(-\qv^T\xv\,,\, q_1\xv + \qv \times \xv)\q^{-1}}\\ 
&= (\qv^T\xv)\qv + q_1(q_1\xv + \qv \times \xv) \nonumber\\
&\quad - (q_1\xv + \qv \times \xv) \times \qv \\
&= (\qv\qv^T + q_1^2\mbf 1 + 2q_1\qv^ \times + \qv^\times\qv^\times) \xv 
\end{align}
Combining the above with \eqref{eq:defQuatMatMaps} yields:
\begin{align}
\label{eq:eulerRodrigues}
& \MH(\q) \\
&= q_1^2 \qid + 2q_1 \qv^\times + {\qv^\times}^2 + \qv {\qv}^T \\
&={\small\arraycolsep=0.3\arraycolsep\ensuremath{\begin{bmatrix}
 q_1^2 + q_2^2 - q_3^2 - q_4^2 &         2q_2q_3 - 2q_1q_4 &         2q_1q_3 + 2q_2q_4 \\
         2q_1q_4 + 2q_2q_3 & q_1^2 - q_2^2 + q_3^2 - q_4^2 &         2q_3q_4 - 2q_1q_2 \\
         2q_2q_4 - 2q_1q_3 &         2q_1q_2 + 2q_3q_4 & q_1^2 - q_2^2 - q_3^2 + q_4^2 \\ 
\end{bmatrix}}}\\
&\overset{*}= (\underbrace{2 q_1^2 - 1}_{q_1^2 - \qv^2}) \qid + 2 q_1 \qv^\times + 2\qv \qv^T \\
&=\begin{bmatrix}
2 q_1^2 - 1 + 2 q_2^2 &
2 (q_2 q_3 - q_4 q_1) &
2 (q_2 q_4 + q_3 q_1) \\
2 (q_2 q_3 + q_4 q_1) &
2 q_1^2 - 1 + 2 q_3^2&
2 (q_3 q_4 - q_2 q_1) \\
2 (q_2 q_4 - q_3 q_1) &
2 (q_3 q_4 + q_2 q_1) &
2 q_1^2 - 1 + 2 q_4^2
\end{bmatrix}\\
&=\begin{bmatrix}
1 - 2 q_3^2 - 2 q_4^2 &
2 (q_2 q_3 - q_4 q_1) &
2 (q_2 q_4 + q_3 q_1) \\
2 (q_2 q_3 + q_4 q_1) &
1 - 2 q_2^2 - 2 q_4^2 &
2 (q_3 q_4 - q_2 q_1) \\
2 (q_2 q_4 - q_3 q_1) &
2 (q_3 q_4 + q_2 q_1) &
1 - 2 q_2^2 - 2 q_3^2
\end{bmatrix}
\end{align}
* is true for  for $\|\q\| = 1$, because then $\crossMat{\qv}^2 = \qv \qv^T  + (q_1^2 - 1) \qid$

\subsection{C++ library Eigen}\label{app:libEigen}
The fastest way to find out what multiplication Eigen seems to be a small test program.
\begin{lstlisting}
#include <iostream>
#include <Eigen/Geometry>

int main() {
  typedef Eigen::Quaterniond Q;
  Q i(0, 1, 0, 0); // w, x, y, z
  Q j(0, 0, 1, 0);
  std::cout << "(i*j).z()=" 
	<< (i * j).z() << std::endl;
}

\end{lstlisting}
Running this program will output \lstinline{(i*j).z()=1}, 
which proves%
\footnote{Assuming that one of the two multiplication is correctly implemented.}
that $i j=k$ holds for the implemented multiplication, \lstinline{*}.

\subsection{Historic investigation}\label{app:history}
Malcolm D. Shuster opened the introduction of \cite{shuster2008}, \quoteinline{The nature of the quaternion}, with :

\quotebig{%
The quaternion \cite{shuster93} is one of the most important representations of the attitude in spacecraft attitude estimation and control.
\ldots
For a brief historical discussion of the quaternion and other attitude representations with references, see reference \cite[pp. 495 - 498]{shuster93}.
After such a long passage of time, the quaternion should be well understood and free of ambiguities. 
Surprisingly, the truth is different, one of the most important inconsistencies has arisen during the past 30 years.
The two most important confusions concern the order of quaternion multiplication and the nature
of the quaternion ``imaginaries,'' both of which are the subject of this article.
\ldots 
That work (\cite{shuster93}) has been cited very frequently within the astrodynamics community over the past fifteen years, and its formulation seems to have become standard there.}

\noindent What had happened before these 30 years? The answer can also be found in \cite{shuster2008}:

\quotebig{%
Hamilton's approach to quaternions \ldots, seemingly in universal use until the publication of reference \cite{Markley1978} and still in almost universal
use until the publication of reference \cite{shuster93}, which, probably, more than any other
work, has been responsible for the change to the natural order of quaternion
multiplication in spacecraft attitude estimation and control. This was, in fact,
an avowed purpose of the author of reference \cite{shuster93}. But although nearly every
writer on spacecraft attitude is aware now of reference \cite{shuster93}, which is cited 
frequently, he or she may not be aware of the inconsistency of reference \cite{shuster93} with many other works on quaternions.}

\noindent To indicate a clear trend towards generally adopting the suggested natural multiplication order \cite{shuster2008} mentions:

\quotebig{Of the more recent texts, four \ldots follow the conventions of reference \cite{shuster2008} and cite it, and one \cite{kuipers1999quaternions} follows the traditional approach.}

\noindent What he did not mention is that these four were, one of his, and three other publications of aerospace engineers, while 
the one still using Hamilton's multiplication was from applied mathematics.
This is nevertheless interesting and might and related to Shuster's own observation when reflecting about why mathematicians might not have reformed the quaternions in \cite{shuster93}: 
\enquote{The concern of pure mathematics is not in representing physical reality efficiently but in exploring mathematical structures \ldots As engineers, our interest is in ``im-pure'' mathematics, contaminated by the needs of practical application.}

Interestingly the very author of \cite{Markley1978} was referencing \cite{shuster93} as the source of the \quoteinline{natural} multiplication order in \cite{markley2000} and in \cite{markley2003}.

In any case a very important contribution of \cite{shuster2008} and \cite{shuster93} is to give up the concept of physical vectors somehow embedded in the algebraic quaternion structure:
\quotebig{There is no need for the “vectors” of the quaternion space to be also vectors in physical space. 
It is sufficient that there be an isomorphism, as there is between physical vectors and their column-vector representations.}

In \cite[p. 473]{shuster93} the author claims that Hamilton's multiplication cannot be made consistent with homomorphy requirement \eqref{eq:pairHom} and his equation (157), which is precisely $\MS$.
An thus \cite{battin1987} would need to change this last relation.
What he apparently missed was that \cite{battin1987} was discussing quaternions used from body to world.
This indicates that at the time Shuster was not fully aware of the big picture described in \prettyref{sec:big-picture-comparison}.
The question is of course where this equation (157) comes from.
The text does not clearly tell but indicates that it is actually derived from the Rodrigues formula and the assumed conversion form rotation vector to quaternion (159), which converts active rotation vectors to active quaternions or passive to passive.
An alternative source could be \cite[8]{Whittaker}.
}

\bibliographystyle{IEEEtran}
\bibliography{quatMult}
\end{document}